\documentclass[lettersize,journal]{IEEEtran}
\usepackage{amsmath,amsfonts}
\usepackage{array}
\usepackage[caption=false,font=normalsize,labelfont=sf,textfont=sf]{subfig}
\usepackage{textcomp}
\usepackage{stfloats}
\usepackage{url}
\usepackage{verbatim}
\usepackage{graphicx}
\usepackage{cite}
\usepackage{bm}
\usepackage{ulem}

\usepackage{multirow}
\usepackage{booktabs}
\usepackage{array}
\usepackage{amsthm}
\usepackage{amsmath}
\usepackage{threeparttable}
\usepackage{makecell}
\usepackage{amssymb}

\usepackage{amsfonts}
\usepackage{amsmath}
\usepackage{graphicx}
\usepackage{textcomp}
\usepackage{xcolor}
\usepackage{rotating}
\usepackage{booktabs} 
\usepackage{multirow}
\usepackage[hang]{footmisc}
\usepackage{caption}
\usepackage{longtable}

\usepackage{algorithm}
\usepackage{algorithmicx}
\usepackage{algpseudocode}
\usepackage[thicklines]{cancel}

\usepackage[hidelinks]{hyperref}

\begin{document}

\title{Cognitive Evolutionary Learning to Select Feature Interactions for Recommender Systems}

\author{
	Runlong~Yu,~\IEEEmembership{Member,~IEEE},
	Qixiang~Shao,
	Qi~Liu,~\IEEEmembership{Member,~IEEE},
	Huan Liu,~\IEEEmembership{Fellow,~IEEE}, \\
	 and~Enhong~Chen,~\IEEEmembership{Fellow,~IEEE}
	 
%
%
%
}

\markboth{Journal of \LaTeX\ Class Files,~Vol.~14, No.~8, August~2022}%
{Shell \MakeLowercase{\textit{et al.}}: A Sample Article Using IEEEtran.cls for IEEE Journals}


\maketitle

\begin{abstract}

Feature interaction selection is a fundamental problem in commercial recommender systems. Most approaches equally enumerate all features and interactions by the same pre-defined operation under expert guidance. Their recommendation is unsatisfactory sometimes due to the following issues: (1)~They cannot ensure the learning abilities of models because their architectures are poorly adaptable to tasks and data; (2)~Useless features and interactions can bring unnecessary noise and complicate the training process. In this paper, we aim to adaptively evolve the model to select appropriate operations, features, and interactions under task guidance. Inspired by the evolution and functioning of natural organisms, we propose a novel \textsl{Cognitive EvoLutionary Learning (CELL)} framework, where cognitive ability refers to a property of organisms that allows them to react and survive in diverse environments. It consists of three stages, i.e., DNA search, genome search, and model functioning. Specifically, if we regard the relationship between models and tasks as the relationship between organisms and natural environments, interactions of feature pairs can be analogous to double-stranded DNA, of which relevant features and interactions can be analogous to genomes. Along this line, we diagnose the fitness of the model on operations, features, and interactions to simulate the survival rates of organisms for natural selection. We show that CELL can adaptively evolve into different models for different tasks and data, which enables practitioners to access off-the-shelf models. Extensive experiments on four real-world datasets demonstrate that CELL significantly outperforms state-of-the-art baselines. Also, we conduct synthetic experiments to ascertain that CELL can consistently discover the pre-defined interaction patterns for feature pairs.

\end{abstract}


\section{Introduction}
\IEEEPARstart{A}{ccurate} targeting of commercial recommender systems is of great importance, in which feature interaction selection plays a key role. Nowadays, feature interaction selection facilitates a variety of applications. For example, with multi-billion dollar business on display advertising, Click-Through Rate (CTR) prediction has received growing interest from both academia and industry communities~\cite{guo2021dual,li2021dual,shi2020deep,lyu2020deep}. Another emerging application is client identifying in FinTech, i.e., intelligent financial advisors of online bank apps attempt to predict whether users are willing to purchase their recommended portfolios~\cite{shao2021toward}. To support such services, it is necessary to well organize the massive high-dimensional sparse user features. Since research has shown that interaction of feature pairs can provide predictive abilities beyond what those features can provide individually~\cite{yu2021xcrossnet}, this brings out the fundamental research problem of feature interaction selection. 

Dash and Liu~\cite{dash1997feature} proposed a general feature selection framework that consists of four steps, that is: 1) generation strategy; 2) evaluation criteria; 3) stopping condition; 4) result validation. 
Based on the framework, researchers developed various feature selection methods which usually fall into three categories: 1) filter; 2) wrapper; 3) embedded. The disadvantages of filters and wrappers lie in poor robustness, besides wrappers are inefficient and unsuitable for large-scale datasets or high-dimensional data. In contrast, embedded methods have become more popular because of their reduced computational demands and less overfitting problems~\cite{li2022survey}. 
Expert-designed operations to model interactions of feature pairs have always dominated extant embedded methods, e.g., factorization machines~(FM)~\cite{rendle2010fm} embed each feature into low-dimensional vector and model interactions by the inner product. These shallow models have limited representation capabilities, which inspires implicit deep learning models to improve over FM~\cite{he2017neural,xiao2017attentional,guo2017deepfm}. 
However, these approaches present the same pre-designed modeling architecture for different tasks or datasets, which leads to a limitation that their architectures are poorly adaptable to tasks and data.
Besides, previous methods simply enumerate all features and interactions by equal relevance, among which useless features and interactions may bring unnecessary noise and complicate the training process. 

To this end, we argue that how to adaptively model task-friendly interactions and quantitatively select relevant features and interactions has become a less-studied urgent problem. 
We expect an ideal feature interaction selection approach should satisfy two desirable abilities: (1) It can evolve to find a proper operation to use for modeling each interaction; (2)~It should distinguish which features and interactions to what extent are relevant to the task.

One way to implement such an evolution of a model is evolutionary learning~\cite{xue2015survey,zhou2019evolutionary,yu2023cognitive}. 
Evolutionary learning refers to a class of evolutionary algorithms that solve complicated search problems in machine learning~\cite{telikani2021evolutionary}, which is naturally fit for the feature selection framework~\cite{tran2017new, cheng2021steering}.
Extant evolution-based feature selection methods perform filters and wrappers~\cite{xue2015survey}, which are not practical for massive high-dimensional commercial data. In contrast, we propose an evolution-based embedded method.
For the concerned feature interaction selection problem, if we regard the relationship between models and tasks as the relationship between organisms and natural environments, interactions of feature pairs can be analogous to double-stranded deoxyribonucleic acid (DNA), of which relevant features and interactions can be analogous to genomes. It is easy to understand that, different traits confer different rates of survival and fitness. The same is true for the selection of features and interactions. A key challenge to accomplishing such evolutionary learning is evaluating the fitness of the model. To cope with the challenge, we propose a fitness diagnosis technique that can reveal the learning abilities of models during training. Compared with previous fitness evaluation techniques, which commonly adopt numerical values to represent the fitness of models, fitness diagnosis can deeply diagnose the abilities of inside components of the model. By doing so, an evolution path can be planned and visualized, thereby enhancing the interpretability of how the model selects operations, features, and interactions that suit the task better.

Along this line, this paper presents a \textsl{Cognitive EvoLutionary Learning (CELL)} framework for feature interaction selection, where cognitive ability refers to a property of organisms that allows them to react and survive in diverse environments.
The CELL framework consists of three stages of learning, i.e., DNA search, genome search, and model functioning. We summarize the three stages in the following:
\begin{enumerate}
	\item \textbf{Stage I: DNA search.}
	We regard the relationship between features and operations as the relationship between nucleotides and linkages. To explore the fittest operation that generates a task-friendly interaction of each feature pair, we extend the operation set with several types of operations as the search space, which is like linkage rules to bind nucleotides. We diagnose the fitness of the model on operations to simulate the survival rates of organisms for natural selection. To tackle the evaluation-consuming challenge of discrete selection, we relax the search space to be continuous so that operation fitness can be optimized by gradient descent. For each feature pair, we retain the fittest operation to model the interaction.

	\item \textbf{Stage II: genome search.}
	Since the genome is a small fraction of DNA that contains genetic information and can influence the phenotype of an organism, the genome search discards or weakens some features and interactions contributing little, to prevent extra noise. To discriminate the relevance of features and interactions, we diagnose the fitness of the model on features and interactions. The search space is the relevance fitness parameters represented by real values, indicating the contributions of features and interactions to the task. Mutation, as the source of genetic variations, occurs when the relevance fitness of an interaction drops to a threshold, resulting in the operation of the interaction mutating into other operations. The mutation mechanism can avoid the search results based on DNA search reaching a suboptimum and benefit genetic diversity.

	\item \textbf{Stage III: model functioning.}
	The model functioning stage utilizes selected relevant features and interactions to capture non-linear interactions further, which is like decoding genes. In this stage, we concatenate retained features and interactions as vectors and then feed them into multilayer perceptrons (MLP), while keeping the relevance fitness as attention units. 
\end{enumerate}

We perform extensive experiments on four datasets. Three of them are publicly available advertising datasets for CTR prediction, and the last one is a financial dataset collected from a high-tech bank for identifying clients. 
To enhance the interpretability of how CELL adaptively selects operations, features, and interactions under task guidance, we visualize the evolution paths of DNA search and genome search. 
 Furthermore, we also conduct synthetic experiments to ascertain that CELL can consistently diagnose to discover the relevant interactions and the pre-defined interaction patterns. 

To summarize, we make the following contributions:

\begin{enumerate}
	\item We propose a nature-inspired feature interaction selection approach named CELL. It can adaptively evolve into different models for different tasks and data, so as to build the architecture under task guidance.
	
	\item We propose a fitness diagnosis technique that can quantitatively analyze and reveal the learning abilities of models during training. It deeply diagnoses the abilities of inside components of models and interprets the mechanism of interaction modeling and selection. 
	
	\item We use three publicly available advertising datasets and a new dataset collected from a high-tech bank to perform experiments. The experimental results show that CELL outperforms the state-of-the-art feature interaction selection models on these tasks. 
\end{enumerate}

\textbf{Overview.} The remainder of this article is organized as follows. In Section~\ref{section2}, we will review some related work of our study. Section~\ref{section3} will introduce the notations, problem definition, and basic operations for generating interactions of feature pairs. Then, the framework of our proposed CELL and the three stages of learning processes will be detailed in Section~\ref{section4}.
  Afterward, Section~\ref{section6} will comprehensively evaluate CELL on real-world datasets and synthetic datasets. Finally, conclusions will be drawn in Section~\ref{section7}.

\section{Related Work} \label{section2}

The related work of our study can be grouped into three categories, namely feature interaction selection, evolutionary learning, and diagnosis techniques. In this section, we summarize the related work as follows.

\subsection{Feature Interaction Selection} 
Since research has shown that interaction of feature pairs can provide predictive abilities beyond what those features can provide individually~\cite{yu2021xcrossnet,xue2015survey}, feature interaction selection based on embedded methods attracts much participation from commercial recommendations, e.g., advertising, client identifying~\cite{liu2012feature,zhou2019deep,xie2021fives,shao2021toward,guo2021embedding}. 
In the earlier time, scholars try to design operations to model interactions in an explicit learning way. For example, factorization machines~(FM)~\cite{rendle2010fm} project each feature into low-dimensional vector and model interactions by the inner product. Field-aware FM (FFM)~\cite{juan2016field} enables each feature to have multiple latent vectors to interact with features from different fields.
However, these models have limited representation capabilities, which inspires implicit deep learning models to improve over FM. For example, Attention FM (AFM)~\cite{xiao2017attentional} and Neural FM (NFM)~\cite{he2017neural} stack deep neural networks on top of the output of the FM to model higher-order interactions. FNN uses FM to pre-train low-order interactions and then feeds embeddings into an MLP~\cite{zhang2016deep}. IPNN (also known as PNN) also uses the interaction results of the FM layer but does not rely on pre-training~\cite{qu2016product,qu2018product}. 
Obviously, these models lack good interpretability. Lian et al. argue that implicit deep learning models focus more on high-order cross features but capture little low-order cross features~\cite{lian2018xdeepfm}. Therefore, recent advances propose a hybrid network structure, namely Wide\&Deep, which combines a shallow component and a deep component with the purpose of learning both memorization and generalization~\cite{cheng2016wide,guo2017deepfm,wang2017deep}. Wide\&Deep framework attracts industry partners from the beginning. As for the first Wide\&Deep model proposed by Google, it combines a linear model and an MLP~\cite{cheng2016wide}. Later on, DeepFM uses an FM layer to replace the shallow part~\cite{guo2017deepfm}. Similarly, Deep\&Cross~\cite{wang2017deep} and xDeepFM~\cite{lian2018xdeepfm} take the outer product of features at the bit- and vector-wise level respectively.
Though achieving some success, most of them follow a manner, which uses the pre-designed modeling architecture and equally enumerates all features and interactions, which suffer from two main problems. 
First, they cannot ensure the learning abilities of models because their architectures are poorly adaptable to tasks and data. Second, useless features and interactions can bring unnecessary noise and complicate the training process. 

\subsection{Evolutionary Learning} 

Evolutionary learning refers to a class of nature-inspired heuristic algorithms that solve complicated search problems in machine learning~\cite{back1996evolutionary,zhou2019evolutionary,juang2021navigation}, which commonly be used in all three parts: preprocessing (e.g., feature selection and resampling), learning (e.g., parameter setting, membership functions, and neural network topology), and postprocessing (e.g., rule optimization, decision tree/support vectors pruning, and ensemble learning)~\cite{tran2017new,zheng2020reconstruction,chen2020evolutionary,song2021fast}. Evolutionary learning also refers to evolutionary AutoML concepts, in which different components of models are automatically determined using evolutionary algorithms, such as architecture and hyperparameters~\cite{telikani2021evolutionary}.
Concepts of evolutionary learning are naturally fit for the feature selection framework~\cite{xue2015survey,tran2017new,cheng2021steering,telikani2021evolutionary}.
Previous evolution-based feature selection methods commonly perform filters and wrappers~\cite{xue2015survey}, which are not practical for massive high-dimensional commercial data. 
Recent approaches have employed AutoML to search neural architectures for CTR prediction~\cite{khawar2020autofeature, song2020towards,liu2020autogroup}. However, they are almost all built on existing work as blocks with complex functionality, such as MLP and FM, where each block is an architecture-level algorithm~\cite{khawar2020autofeature,song2020towards}.
In general, for population-based NAS approaches, large population size and massive generations are usually required to address the huge search space issue, e.g., AutoCTR sets population size as 100, and every individual search runs on a single GPU for days~\cite{song2020towards}.
In contrast to searching coarse-grained upper-level architectures, our work searches fine-grained basic-level operations. To the best of our knowledge, CELL is the first work that utilizes the meta-heuristic mutation mechanism for interaction search. 

\subsection{Diagnosis Techniques} 

In educational psychology, cognitive diagnosis techniques have been developed to access examinees’ skill proficiency and can be roughly divided into two categories: continuous and discrete~\cite{dibello200631a, de2011generalized}. These works follow the Monotonicity Theorem that assumes that examinees’ proficiency increases monotonously with probabilities of giving the right responses to test items~\cite{tong2021item}. Furthermore, recent works leverage diagnosis techniques to reveal the proficiency of trial lawyers in legal fields \cite{yue2021circumstances}. Also, research shows that one can predict the outcome of e-sports matches through the diagnosis of players \cite{gu2021neuralac}. However, previous diagnosis techniques are mostly oriented toward human beings. Our work inherits the idea of both continuous and discrete diagnosis techniques but changes whom to diagnose from examinees to learning models. Noticeable improvement in machine learning benefits from the inspiration of the way of human learning, whereas fitness evaluation of machine learning models comes to a standstill. 
In contrast, our work provides a fitness diagnosis targeting machine learning models to understand their learning status during training.  We diagnose the fitness of the model to simulate the survival rates of organisms for natural selection, thereby an evolution path can be planned and visualized. It interprets the mechanism of interaction modeling and selection.

\section{Preliminaries} \label{section3}

In this section, we state the formal problem definition of feature interaction selection and introduce some basic operations for generating interactions of feature pairs. 

\subsection{Problem Statement.}
We define a general form of feature interaction selection problem. If the dataset consists of $n$ instances $(\pmb{f},y)$, where $\pmb{f}=[f_1,\dots,f_m]$ indicates instance features including $m$ fields, and $y \in \{1,0\}$ indicates an observed user behavior,
the feature interaction selection problem can be defined as how to precisely give the predictive result through the learned model $\hat{y}: \mathcal{M}(\pmb{f}, \pmb{g}(\pmb{f})) \mapsto [0,1]$,  where $\pmb{g}$ denotes the set of operations between feature pairs, and $\pmb{g}(\pmb{f})$ denotes the set of interactions.
Usually, the prediction model $\mathcal{M}$ suffers a Logloss (cross-entropy loss) function~\cite{tao2022sminet}, given as:
\begin{equation}
\mathcal{L} (\mathcal{M}) = - \frac{1}{|B|} \sum_{t\in B} y_t \log(\hat{y_t}) + (1-y_t) \log(1 - \hat{y_t}),
\end{equation} where $B$ denotes the set of instance indices in a mini-batch, $\hat{y}$ denotes the predictive result given through the learned model.

\begin{figure*}[!t]
	\centering
	\includegraphics[width=\linewidth]{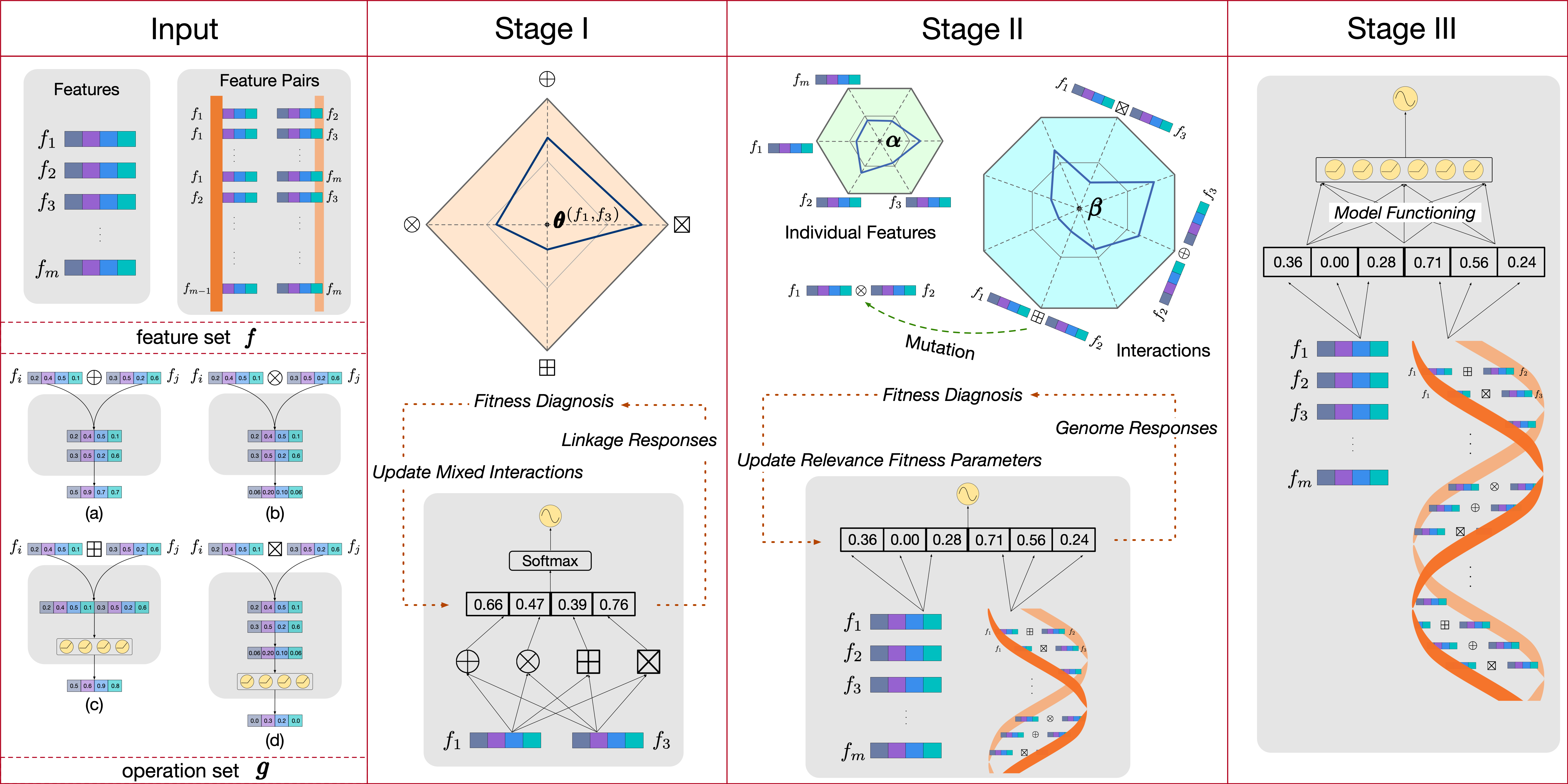}
	\caption{An illustration of the CELL framework. Inputs include the feature set and the operation set. Stage I: DNA search explores different operations to model the interaction of feature pairs. Stage II: genome search discards or weakens some features and interactions contributing little, and remodels some interactions by mutating their operations. Stage III: the model functioning stage utilizes selected relevant features and interactions to capture non-linear interactions further. } 
	\label{fig1} 
\end{figure*}

\subsection{Operations.}

As the fundamental components in feature interaction selection, operations are regarded as functions where two individual features are converted into an interaction. For the sake of simplicity, we adopt four representative operations, i.e., $\pmb{g}=\{ \oplus, \otimes,  \boxplus, \boxtimes \}$, to present an instantiation of CELL, which are highly used in previous work~\cite{khawar2020autofeature, song2020towards,liu2020autogroup}. 
As shown in the operation set part of Fig.~\ref{fig1},  the following operations are available for selection:

\begin{enumerate}
	\item \textbf{Element-wise sum} ($\oplus$): It takes two input vectors of dimension~$|f|$ and outputs a vector of dimension~$|f|$ that contains their element-wise sum. It has no parameters.
	\item \textbf{Element-wise product} ($\otimes$): It takes two input vectors of dimension~$|f|$ and outputs a vector of dimension~$|f|$ that contains their element-wise product. It has no parameters.
	\item \textbf{Concatenation and a feed-forward layer} ($\boxtimes$): It takes two input vectors of dimension $|f|$, concatenates them, and passes them through a feed-forward layer with $\rm{ReLU}$ activation functions to reduce the dimension of the output vector to $|f|$.
	\item \textbf{Element-wise product and a feed-forward layer} ($\boxplus$): It takes two input vectors of dimension $|f|$, passes their element-wise product through a feed-forward layer with $\rm{ReLU}$ activation functions to output a vector of dimension $|f|$.
\end{enumerate} 

The complexities of $\oplus$ and $\otimes$ are both $O(1)$. The complexities of $\boxtimes$ and $\boxplus$ are both $O(|f|)$. In practice, we simultaneously optimize operations $\boxtimes$, $\boxplus$ and feature embedding $\pmb{f}$. Thus, for an interaction of a feature pair, the complexity is $O(|f|)$.

\newtheorem{theorem}{Theorem}[section]
\newtheorem{assumption}{Assumption}[section]
\newtheorem{proposition}{Proposition}[section]

\section{Cognitive Evolutionary Learning} \label{section4}

In this section, we will introduce the framework of CELL, which consists of three stages of learning, i.e., DNA search, genome search, and model functioning. The overall framework is illustrated in Fig. \ref{fig1}. The core idea of CELL is to adaptively model task-friendly interactions and quantitatively select relevant features and interactions so as to build the architecture under task guidance, which is inspired by the evolution and functioning of natural organisms. We consider a feature interaction selection process as a multi-stage iterative search process, which can be viewed as a natural organism striving to evolve better traits for higher rates of fitness. The traits of an organism can be inherited via two aspects, i.e., the structure of DNA and the genome regions of DNA. We diagnose the fitness of the model to simulate the survival rates of organisms for natural selection, which interprets the mechanism of interaction modeling and selection. The three stages are detailed successively as follows.

\subsection{DNA Search}

We regard the relationship between features and operations as the relationship between nucleotides and linkages, so the evolution of DNA is expected to explore the fittest operation that generates a task-friendly interaction of each feature pair. Following various linkages of nucleotides within DNA, we extend the operation set with four types of operations as the search space, i.e.,  $\pmb{g}=\{ \oplus, \otimes,  \boxplus, \boxtimes \}$. If $g_k$ is a candidate operation from the operation set $\pmb{g}$, an interaction $g_k(f_i, f_j)$ is modeled by the operation $g_k$ applied to a feature pair $(f_i, f_j)$. Along this line, the model can be analogous to an organism, of which traits can be affected by the linkage of nucleotides. It responds to instance predictions,  just like the organism responds to natural environments. Our intuitive goal here is to diagnose the fitness of the model on operations through the difference between the linkage responses and true labels, as elaborated below:

Let a set of continuous variables $\pmb{\theta}=\{\pmb{\theta}^{(i,j)} | 1\leqslant i<j\leqslant m \}$ parameterize the fitness of the model on operations. Our diagnosis goal is to learn $\pmb{\theta}^{(i,j)}=\{\theta_{g_k}^{(i,j)} | g_k\in \pmb{g} \}$  for each feature pair $(f_i, f_j)$ of an instance.

For an organism, its fitness increase monotonously with probabilities of the DNA decoding vital traits. Analogously, we assume that the fitness of the model increase monotonously with probabilities of interacting feature pairs by suitable operations. Motivated by this, we put forward that $\pmb{\theta}^{(i,j)}$ satisfies the following property: 
\begin{proposition} 
	$\forall g_k, g_k' \in \pmb{g}$, $\pmb{\theta}^{(i,j)}$ satisfies $\theta_{g_k}^{(i,j)} > \theta_{g_k'}^{(i,j)}$ when $\mathcal{L} (\mathcal{M}(g_k(f_i, f_j)) ) < \mathcal{L} (\mathcal{M}(g_k'(f_i, f_j)) )$. \label{proposition1}
\end{proposition}

For the sake of simplicity, we use $ \mathcal{M}(g_k(f_i, f_j)) $ to denote the learning model that selects the operation $g_k$ to interact the feature pair $(f_i, f_j)$. $\mathcal{L} (\mathcal{M})$ is the suffered Logloss function as we mentioned before. The above proposition provides us with a way to optimize the relative value of the model fitness on operations. There are obvious advantages to implementing DNA search with continuous fitness rather than categorical choices. If we search over discrete sets of interactions modeled by candidate operations, it will require much more evaluation. Contrastively, we relax the categorical choices to be continuous so that the set of operations can be optimized with respect to its performance on the validation set by gradient descent. As opposed to inefficient black-box search, the data efficiency of gradient-based optimization allows DNA search to use much fewer computational resources. 

Inspired by DARTS algorithm~\cite{liu2019darts}, for each feature pair, we adopt a mixed interaction $\bar{g}(f_i, f_j)$ given by a softmax over all possible operations as: 
\begin{equation}
\bar{g}(f_i, f_j) = \sum_{g_k\in \pmb{g}} \frac{\exp\left(\theta_{g_k}^{(i,j)}\right)}{\sum_{g_l\in \pmb{g}} \exp\left(\theta_{g_l}^{(i,j)}\right)} g_k(f_i, f_j),
\end{equation} where the coefficient of candidate operations is called mixed linkage. We define the linkage response as the weighted sum of all feature pairs interacted by mixed linkages: 
\begin{equation}
\hat{y} = \mathcal{M}(\pmb{w} \cdot \pmb{g}(\pmb{f})) = {\rm{Sigmoid}} \bigg(\sum_{1\leqslant i<j\leqslant m} w_{i,j} \cdot \bar{g}(f_i, f_j) \bigg),
\end{equation} where $w_{i,j}$ is the weight parameter of $\bar{g}(f_i, f_j) $. 
If we consider an interaction given by the mixed linkage between a feature pair as a DARTS process, the linkage response of all feature pairs is given by multiple DARTS processes (practically hundreds and thousands of them) that run in parallel. 

According to Proposition \ref{proposition1}, DNA search can optimize $\pmb{\theta}$ via the difference between the linkage responses and true labels. We adopt to jointly optimize $\pmb{\theta}$ and $\pmb{w}, \pmb{f}$ as a bilevel optimization problem.  Hence, our goal  is to find $\pmb{\theta}^*$ that minimize the validation loss $\mathcal{L}_{val}(\pmb{w}^*, \pmb{f}^*, \pmb{\theta}^* ) $, where $\pmb{w}^*, \pmb{f}^*$ associated with the operations are obtained by minimizing the training loss $\pmb{w}^*, \pmb{f}^* = \arg \min_{\pmb{w}, \pmb{f}} \mathcal{L}_{train}(\pmb{w}, \pmb{f}, \pmb{\theta}^*)$. This implies that $\pmb{\theta}$ is the upper-level variable and $\pmb{w}, \pmb{f}$ are the lower-level variables:
\begin{equation}  
\begin{aligned}
&\min_{\pmb{\theta}} \mathcal{L}_{val}\left( \pmb{w}^* (\pmb{\theta}), \pmb{f}^*, \pmb{\theta}  \right), \\
{\rm {s.t.}}  \quad & \pmb{w}^* (\pmb{\theta}), \pmb{f}^* = \arg \min_{\pmb{w}, \pmb{f}} \mathcal{L}_{train}(\pmb{w}, \pmb{f}, \pmb{\theta}).
\end{aligned} \label{eq2} 
\end{equation} To reduce the expensive inner optimization in Eq. (\ref{eq2}), we can use a simple approximation that assumes the current $\pmb{w}, \pmb{f}$ are the same as $\pmb{w}^*, \pmb{f}^*$. Related techniques have been used in DARTS~\cite{liu2019darts}, meta-learning for model transfer~\cite{finn2017model}, gradient-based hyperparameter tuning~\cite{luketina2016scalable} and unrolled generative adversarial networks~\cite{metz2016unrolled}. 

Through the approximation, we can replace $\pmb{w}^* (\pmb{\theta}), \pmb{f}^*$ by adapting $\pmb{w}, \pmb{f}$ using only a single training step, without solving the inner optimization completely by training until convergence, given as: 
\begin{equation}
\begin{aligned}
\nabla_{\pmb{\theta}} \mathcal{L}_{val}\left( \pmb{w}^* (\pmb{\theta}), \pmb{f}^*, \pmb{\theta}  \right) \approx &  \nabla_{\pmb{\theta}} \mathcal{L}_{v a l}\big( \pmb{w}-\xi \nabla_{\pmb{w}} \mathcal{L}_{train}(\pmb{w}, \pmb{f},\pmb{\theta}), \\ & \pmb{f}-\xi \nabla_{\pmb{f}} \mathcal{L}_{train}(\pmb{w}, \pmb{f},\pmb{\theta}),\pmb{\theta} \big ),
\end{aligned}
\label{eq4}
\end{equation} where $\xi$ is the learning rate. Eq. (\ref{eq4}) requires only a forward pass for $\pmb{w}, \pmb{f}$ and a backward pass for $\pmb{\theta}$, which is much more efficient.
At the end of DNA search, we replace $\bar{g}(f_i, f_j)$ with the interaction modeled by the fittest operation and obtain a discrete choice of operation, i.e., $g = \arg \max_{g_k\in \pmb{g}} \theta_{g_k}^{(i,j)}$.

The overall time complexity of DNA search is practicable owing to the real-parameter model fitness mechanism and the approximation method. Compared to searching over discrete sets of candidate operations, the complexity is reduced from $ O (  |\pmb{w}| + 4m \cdot |\pmb{f}|   \cdot| \pmb{\theta}| ) $ to $ O(  |\pmb{w}| + 4m \cdot|\pmb{f}| + | \pmb{\theta}| )  $, where $|\pmb{w}|=m\cdot (m-1)/2$, $|\pmb{f}| = m \cdot |f|$, and $| \pmb{\theta}| = 2m\cdot (m-1)$.  
Note that Eq.~(\ref{eq4}) will reduce to $ \nabla_{\pmb{\theta}} \mathcal{L}_{v a l}( \pmb{w}, \pmb{f}, \pmb{\theta}  )$ if $\pmb{w}, \pmb{f}$ are already local optima, because $\nabla_{\pmb{w}} \mathcal{L}_{train}(\pmb{w}, \pmb{f},\pmb{\theta})=0$ and $\nabla_{\pmb{f}} \mathcal{L}_{train}(\pmb{w}, \pmb{f},\pmb{\theta})=0$.

\subsection{Genome Search}

DNA search provides us a way to select appropriate operations from the operation set to model interactions of feature pairs. Regarding the model as an organism, in DNA search, the nucleotides evolve to be bound by linkages so that nucleotide chains become a double helix structure. Since the genome is a small fraction of DNA that contains genetic information and can influence the phenotype of an organism, our next goal is to find genome regions in DNA. 

For an organism, if the regions in DNA decodes a phenotype that benefits survival, it will have better fitness; if the phenotype decoded by the regions does not benefit survival, it will have worse fitness. The evolutionary strategy should allow the former (i.e., the genome) to retain phenotype genetic information, which inspires us to formulate the importance of features and interactions with the relevance fitness parameters. Therefore, our intuitive goal here is to discriminate the relevance of features and interactions, so as to enhance the relevant features and interactions, meanwhile, discard or weaken some features and interactions contributing little. 

Let $\pmb{\alpha} = \{ \alpha_i | 1\leqslant i\leqslant m\} $ and $\pmb{\beta} =\{ \beta_{i,j} | 1\leqslant i<j\leqslant m \} $ denote the relevance fitness parameters of features $\pmb{f} $ and interactions $\pmb{g(f)} $. 
Our diagnosis goal is to learn $\alpha_i$ for each feature $f_i$ and $\beta_{i,j}$ for each interaction $g(f_i, f_j)$ of an instance. We put forward that $\pmb{\alpha}, \pmb{\beta} $ satisfy the following property: 
\begin{proposition} 
	Let $ \pmb{\alpha}, \pmb{\beta} = \arg \min_{\pmb{\alpha}, \pmb{\beta}} \mathcal{L} (\mathcal{M})  $. $ \pmb{\alpha} $ satisfies $|\alpha_i|  > |\alpha_j|$ when $ \mathcal{L} (\mathcal{M} (f_i, w/o(f_j))) < \mathcal{L} (\mathcal{M} (w/o(f_i), f_j ) ) $, and $ \pmb{\beta} $ satisfies $|\beta_{i,j}| > |\beta_{k,l}| $ when $ \mathcal{L} (\mathcal{M} (g(f_i, f_j), w/o(g(f_k, f_l))) ) < \mathcal{L} (\mathcal{M} ( w/o(g(f_i, f_j)), g(f_k, f_l)) ) $.   \label{proposition2} 
\end{proposition}

For the sake of simplicity, in the above proposition, we use $ \mathcal{L} (\mathcal{M} (f_i, w/o(f_j))) $ to denote a comparative form of the learning model that selects feature $f_i$ and discards feature $f_j$. Analogously, $ \mathcal{L} (\mathcal{M} (g(f_i, f_j),w/o(g(f_k, f_l) )) ) $ denotes the learning model that selects interaction $g(f_i, f_j) $ and discards feature $g(f_k, f_l)$. Indeed, the core idea of fitness diagnosis in genome search can be viewed as a comparative survival scheme, which enhances the relevant features and interactions but discards or weakens others through the difference between the genome responses and true labels. 

Along this line, the genome response of the current learning model can be given as follows: $\hat{y} = \mathcal{M}( \pmb{\alpha} \cdot \pmb{f} ,  \pmb{\beta} \cdot \pmb{g}(\pmb{f})) 
= {\rm{Sigmoid}} \left( \sum_{i=1}^{m} \alpha_{i}  \cdot f_i +  \sum_{1\leqslant i<j\leqslant m} \beta_{i,j}  \cdot g(f_i, f_j) \right)$, where the $\alpha_i$ is the relevance fitness parameter of feature $f_i$, and $\beta_{i,j}$ is the relevance fitness parameter of interaction $g(f_i, f_j)$. 
We propose to optimize $\pmb{\alpha}$, $\pmb{\beta} $ simultaneously with feature embeddings. Specifically, feature embeddings are learned by Adam optimizer~\cite{kingma2014adam}, while $\pmb{\alpha}$, $\pmb{\beta} $ are learned by regularized dual averaging~(RDA) optimizer~\cite{xiao2009dual, chao2019generalization}.
RDA optimizer is an online optimization method aiming at getting sparse solutions. The reason why we can guarantee sparse solutions of $\pmb{\alpha}$, $\pmb{\beta}$ is because of the truncation mechanism of the RDA optimizer. When the absolute value of the cumulative gradient average value in a certain dimension is less than a threshold, the weight of that dimension will be set to 0, resulting in the sparsity of the relevance fitness parameters~\cite{xiao2009dual,liu2020autofis}. We update $\pmb{\alpha}$, $\pmb{\beta} $ at each gradient step $t$ with data $B_t$ as: 
\begin{equation}
\pmb{\alpha}_{t+1}, \pmb{\beta}_{t+1} =  \mathcal{S}_{h(t, \gamma)} \bigg \{ (\pmb{\alpha}_0, \pmb{\beta}_0) - \gamma \sum_{i=0}^{t} \nabla \mathcal{L}(\pmb{\alpha}_i, \pmb{\beta}_{i}; B_i) \bigg \},
\end{equation}where $\mathcal{S}_h: v \mapsto \text{sign}(v) \cdot \max\{|v| - h, 0\} $ is the soft-thresholding operator, $\pmb{\alpha}_0, \pmb{\beta}_0$ are initializers chosen at random, $\gamma$ is the learning rate, $h(t, \gamma)=c \gamma^{1/2}(t \gamma)^{\mu}$ is the tuning function, $c$ and $\mu$ are adjustable hyper-parameters as a trade-off between accuracy and sparsity. 
To avoid the expensive inner optimization of the gradient of $\pmb{\alpha}$, $\pmb{\beta} $ and feature embeddings, 
the parameters are updated together using one-level optimization with gradient descent on the training set by descending on $\pmb{\alpha}$, $\pmb{\beta}$ and $\pmb{f}$ based on:
\begin{equation}
\nabla_{\pmb{f}} \mathcal{L}_{train}(\pmb{f}_{t-1}, \pmb{\alpha}_{t-1}, \pmb{\beta}_{t-1})
\end{equation}
\begin{equation}
\text{and}  \;\; \nabla_{\pmb{\alpha,\beta}} \mathcal{L}_{train}(\pmb{f}_{t-1}, \pmb{\alpha}_{t-1}, \pmb{\beta}_{t-1}).
\end{equation} In this setting, $\pmb{\alpha}$, $\pmb{\beta}$ and $\pmb{f}$ can explore their search space freely until convergence. The complexity is $O(m\cdot |\pmb{f}| + |\pmb{\alpha}|+|\pmb{\beta}| )$, where $|\pmb{f}| = m \cdot |f|$, $|\pmb{\alpha}| = m$, and $|\pmb{\beta}| = m\cdot (m-1)/2 $. 

\begin{algorithm} [!t]
	\caption{Cognitive Evolutionary Learning} \label{alg1}
	\textbf{Input}: Training dataset of $n$ instances $(\pmb{f},y)$, operation set  $\pmb{g}$.
	\begin{algorithmic}[1] 
		\State Randomly initialize fitness parameters $\pmb{\theta}, \pmb{\alpha}, \pmb{\beta}$.
		\Procedure{DNA Search}{$\pmb{\theta}$} 
		\While{not converged}
		\State Update model fitness $\pmb{\theta}$ by descending $ \nabla_{\pmb{\theta}} \mathcal{L}_{v a l}\big( \pmb{w}-\xi \nabla_{\pmb{w}} \mathcal{L}_{train}(\pmb{w}, \pmb{f},\pmb{\theta}), \pmb{f}-\xi \nabla_{\pmb{f}} \mathcal{L}_{train}(\pmb{w}, \pmb{f},\pmb{\theta}),\pmb{\theta} \big )$
		\State Update weights $\pmb{w}$ and feature embeddings $\pmb{f}$ by descending $\nabla_{\pmb{w}} \mathcal{L}_{train}(\pmb{w}, \pmb{f},\pmb{\theta}), \nabla_{\pmb{f}} \mathcal{L}_{train}(\pmb{w}, \pmb{f},\pmb{\theta})$
		\EndWhile
		\State \Return the selected operations $g = \arg \max_{g_k\in \pmb{g}} \theta_{g_k}^{(i,j)}$
		\EndProcedure
		\item[]
		\Procedure{Genome Search}{$\pmb{\alpha}, \pmb{\beta}$} 
		\While{not converged}
		\State Update the relevance fitness $\pmb{\alpha}, \pmb{\beta}$ by descending $ \nabla_{\pmb{\alpha,\beta}} \mathcal{L}_{train}(\pmb{f}_{t-1}, \pmb{\alpha}_{t-1}, \pmb{\beta}_{t-1}) $ \Comment{Use RDA optimizer}
		\State Update feature embeddings $\pmb{f}$ by descending $\nabla_{\pmb{f}} \mathcal{L}_{train}(\pmb{f}_{t-1}, \pmb{\alpha}_{t-1}, \pmb{\beta}_{t-1}) $
		\If{$\beta_{i,j} < \lambda$ for every $\tau$ steps}
		\State  $g_k$ of $g_k(f_i, f_j)$ probabilistically mutates
		\EndIf
		\EndWhile
		\State \Return the relevance fitness parameters $\pmb{\alpha}, \pmb{\beta}$
		\EndProcedure
		\item[]
		\Procedure{Model Functioning}{$\pmb{\alpha}, \pmb{\beta}$} 
		\While{not converged}
		\State Update weights $\pmb{w}$ of the MLP and $\pmb{f}$ by descending $\nabla_{\pmb{w}} \mathcal{L}_{train}(\pmb{w}, \pmb{f}), \nabla_{\pmb{f}} \mathcal{L}_{train}(\pmb{w}, \pmb{f})$
		\EndWhile
		\State \textbf{return} the final predictive model
		\EndProcedure
	\end{algorithmic}
\end{algorithm}

To further explore interactions modeled by operations between features, we propose a mutation mechanism, inspired by the source of genetic variations. Mutation probabilistically occurs when the relevance fitness drops to a threshold, resulting in the operation of the interaction mutating into other operations. In general, the mutation mechanism can avoid the search result based on DNA search reaching a suboptimum and benefit the genetic diversity. 

When $\beta_{i,j}$ drops to a threshold $\lambda$ for every $\tau$ steps, the mutation is applied with probability $1 / \sigma$, which means that, to regenerate a new interaction, the operation $g_k$ of the interaction $g_k(f_i, f_j)$ mutates into another operation $g_l $, given as: 
\begin{equation}
g_k = \left\{
\begin{aligned}
g_l  \;  \text{with probability} \, 1 / \sigma, \;\;\;\;  & \textbf{if} \quad  \beta_{i,j} < \lambda, \\
g_k, \;\;\;\;\;\;\;\;\;\;\;\;\;\;\;\;\;\; & \textbf{otherwise}.
\end{aligned}
\right.
\end{equation}
where  $g_l$ is randomly selected from the operation set as $g_l = \{ g\, | \, g \in \pmb{g}, g \neq g_k \}$. After the mutation, the new interaction $g_l(f_i, f_j)$ replaces the old irrelevant interaction $g_k(f_i, f_j)$ and participates in next $\tau$ steps of genome search. 

In certain sense, genome search makes a good balance between exploitation and exploration of interactions, because it enhances relevant features and interactions, meanwhile it remodels irrelevant interactions by mutating their operations.

\subsection{Model Functioning}

Through repeated replication and transcription, the genetic information of DNA translates to make matching protein sequences. In this way, various possible functions could be proposed for an organism. To simulate this natural process, we retrain the model. Relevant features and interactions are selected according to their relevance fitness parameters $\pmb{\alpha}$, $\pmb{\beta}$. 
If $\alpha_{i}=0$ or $\beta_{i,j}=0$, the corresponding individual features or interactions are fixed to be discarded permanently. To further capture non-linear interactions with selected relevant features and interactions, in the model functioning stage, we concatenate features and interactions as vectors and feed them into an MLP: 
\begin{equation}
\begin{aligned}
\hat{y} = & {\rm{Sigmoid}} ( {\rm{MLP}} ( [\alpha_1  \cdot f_1, \dots, \alpha_m \cdot f_m, \beta_{1,2} \cdot g(f_1,f_2), \\ 
& \dots, \beta_{m-1, m} \cdot g(f_{m-1}, f_m)] ) ),
\end{aligned}
\end{equation} where the relevance fitness parameters $\pmb{\alpha}$, $\pmb{\beta}$ are fixed and serve as attention units. In this stage,  weights $\pmb{w}$ of the MLP and $\pmb{f}$ are learned by Adam optimizer~\cite{kingma2014adam}.
The time complexity depends on the number of neurons per layer and the depth of layers of the MLP.

Algorithm \ref{alg1} outlines the procedures of the CELL framework. DNA search optimizes the model fitness on operations and returns the selected fittest operations for interacting each feature pair. Genome search optimizes the relevance fitness parameters and mutates the operation when the relevance fitness of the interaction drops to a threshold. The model functioning procedure retrains the model based on selected relevant features and interactions to capture non-linear interactions. 

\textbf{Discussion on CELL.} 
We view CELL as a novel nature-inspired framework for feature interaction selection, which embodies three tribes of artificial intelligence algorithms~\cite{domingos2015master}. First, from the perspective of the analogizer algorithm, all of the fine-grained operations in CELL follow the analogized patterns, which embed the features into low-dimensional spaces to infer their similarities. Second, from the perspective of the evolutionary algorithm, the search procedures in CELL follow the evolution process, that is, the parent model generates the offspring model based on fitness. Furthermore, CELL combines the advantages of gradient descent search and stochastic search, where the former is efficient and does not require massive processing units, while the latter can explore the search space of interactions to avoid reaching a suboptimum. Moreover, compared with previous fitness evaluation, which commonly adopts numerical values to represent the fitness of models, CELL deeply diagnoses the abilities of inside components of the model, which can enhance interpretability and can help discover order through the diagnosis results. Third, from the perspective of the connectionist algorithm, the model functioning procedure calculates the nonlinear connections among features and interactions to instance prediction. The border crossing, where different basic algorithms meet, has always been one of the most promising parts of the research for next-generation learning machines. CELL is more close to this goal and is expected to facilitate feature interaction selection better.

In practice, one may concern about the scalability of CELL when the operation set is expanded. Suppose the operation set has $|\pmb{g}|$ operations, because only DNA search is affected by $\pmb{g}$, its time complexity varies with the number of operations as $ O( |\pmb{g}| \cdot m^2 \cdot|f| )$. Therefore, the time complexity varies significantly with the feature fields and not much with the size of the operation set. This shows that our method is highly scalable and practical on a larger set of operations. For industrial use cases where hundreds and thousands of feature fields are available, common embedded methods are always incapable of solution, and so is our proposed CELL. In these cases, we usually need to preprocess the features with filter methods in advance, and remove the evidently irrelevant features first. After that, the computational cost is workable for CELL. In the experimental section, we will present an example where we adopt this strategy to deploy CELL to an industrial FinTech scenario.

\section{Experiments} \label{section6}

In this section, we conduct comprehensive experiments to demonstrate the effectiveness of CELL. Specifically, we first describe the datasets and experimental settings used in the experiments. Then, we compare the recommendation performance of CELL with baseline approaches in terms of AUC and Logloss. After that, we visualize the evolution paths of DNA search and genome search to show the interpretability of CELL. Finally, we also conduct synthetic experiments to ascertain that CELL can consistently diagnose to discover the relevant interactions and the pre-defined interaction patterns. All experiments are conducted on a Linux server with an NVIDIA Tesla K80 GPU.

\begin{table}[t] 
	\caption{The statistics of the four datasets.} 
	\begin{tabular}{p{1cm} <{\centering}  p{1.6cm}<{\centering}  p{1.6cm}<{\centering}   p{0.7cm}<{\centering}  p{1.6cm}<{\centering} }
		\toprule
		Dataset & \#Instances & \#Categories & \#Fields & Positive ratio \\
		\midrule
		Criteo & 45,840,617  & 34,290,882  & 39 & 0.34    \\
		Avazu & 40,428,967  & 9,449,445   & 23 & 0.20    \\
		Huawei     & 41,907,133 & 1,096,074  & 35 & 0.04  \\
		FinTech & 31,182,310 & 7,037,326  & 36 & 0.05 \\
		\bottomrule
	\end{tabular}
	\label{table01}
	\vspace{-0.1cm}
\end{table}

	\vspace{-0.1cm}
\subsection{Datasets} 

We use four datasets in the experiments. The statistics are reported in Table \ref{table01}. Three of them are publicly available advertising datasets for CTR prediction, i.e., 
Criteo\footnote{https://www.kaggle.com/c/criteo-display-ad-challenge/data}, Avazu\footnote{https://www.kaggle.com/c/avazu-ctr-prediction/data}, Huawei\footnote{https://www.kaggle.com/louischen7/2020-digix-advertisement-ctr-prediction}.
We describe the advertising datasets and the pre-processing steps below.
\begin{enumerate}
	\item \textbf{Criteo} is a famous benchmark dataset for CTR prediction published by Criteo AI Lab. Criteo contains one month of click logs with billion of data, and a small subset of it was published in Criteo Display Advertising Challenge in 2013. We select ``day~6-12" for training and evaluation. 13~numerical fields are converted into one-hot features through bucketing, where the features in a certain field appearing less than 20 times are set as a dummy feature ``other".
	\item \textbf{Avazu} was published in the Avazu Click-Through Rate Prediction contest in 2014. This dataset contains users' mobile behaviors including whether a displayed mobile ad is clicked by a user or not. It has 23 feature fields spanning from user/device features to ad attributes. We select 10 days of data for training and evaluation.  
	\item \textbf{Huawei} was published in Huawei DIGIX Advertisement CTR Prediction in 2020. The datasets contain the advertising behavior data collected from seven consecutive days.  It has 35 feature fields spanning from user/device features to ad attributes.  
\end{enumerate} 

The last dataset is a financial dataset for client identifying, i.e., intelligent financial advisors of online bank apps predict whether their users are willing to purchase recommended portfolios. As best as we know, there are no such public datasets currently. To protect data privacy, all the user information has been desensitized and the feature fields have been anonymized. The details of it are given below.
\begin{enumerate}\setcounter{enumi}{3}
	\item \textbf{FinTech} is collected from a high-tech bank in China and ranges from January 1, 2020, to December 1, 2021. It totally has 2257 feature fields on account of the wide range of businesses involved. As many features are evidently irrelevant to the task, we filter them by preliminary screening on a predefined metric, i.e., Information Value (IV). The $IV$ of a feature field can be calculated according to $IV = \sum_{i=1}^n (Pos_i - Neg_i) \times \ln (\frac{Pos_i}{Neg_i})$, where $Pos_i$ and $Neg_i$ respectively represent of the proportion of positive and negative instances under the corresponding features. The higher $IV$ of a feature field is, the better predictive ability the feature has. After the preliminary screening, we retain 600 feature fields. To further have a fine selection, we use genome search to select relevant features and retain 36 feature fields to construct this FinTech dataset, which contains user basic features, e.g., ages and genders, user financial features, e.g., asset conditions and risk appetites, and user behavioral features, e.g., investments and consumptions. 
\end{enumerate}

\newcolumntype{L}[1]{>{\raggedright\arraybackslash}p{#1}}
\newcolumntype{C}[1]{>{\centering\arraybackslash}p{#1}}
\newcolumntype{R}[1]{>{\raggedleft\arraybackslash}p{#1}}

\begin{table*}[!t]\footnotesize
	\centering
	\caption{ Performance comparison. Impr is the relative AUC improvement. $^{\star}$ and $^{\lozenge}$ represent significance level $p$-value $<10^{-5}$ and $p$-value $<0.05$ of comparing CELL with the best baseline (indicated by underlined numbers).}
	\begin{tabular}{L{2.5cm}C{0.80cm}C{0.80cm}C{0.90cm}C{0.80cm}C{0.80cm}C{0.90cm}C{0.80cm}C{0.80cm}C{0.90cm}C{0.80cm}C{0.80cm}C{0.90cm}}
		\toprule
		\multicolumn{1}{l}{\multirow{2}[4]{*}{\small{Model Name}}} & \multicolumn{3}{c}{\small{Criteo}} & \multicolumn{3}{c}{\small{Avazu}} & \multicolumn{3}{c}{\small{Huawei}} & \multicolumn{3}{c}{\small{FinTech}} \\
		\cmidrule(lr){2-4}   \cmidrule(lr){5-7}   \cmidrule(lr){8-10}   \cmidrule(lr){11-13}    & AUC   & Logloss & Impr(\%) & AUC   & Logloss & Impr(\%) & AUC   & Logloss & Impr(\%) & AUC   & Logloss & Impr(\%) \\
		\midrule
		LR & 0.7784 & 0.4692 & 4.37  & 0.7627 & 0.3896 & 5.02  & 0.7658 & 0.1322 & 4.32 & 0.8477 & 0.1539 & 2.50 \\
		FM   & 0.7940 & 0.4583 & 2.32  & 0.7895 & 0.3746 & 1.46  & 0.7860 & 0.1281 & 1.64 & 0.8511 & 0.1533 & 2.09 \\
		AFM & 0.7985 & 0.4520 & 1.74  & 0.7867 & 0.3759 & 1.82  & 0.7925 & 0.1259 & 0.81 & 0.8569 & 0.1505 & 1.40 \\
		FFM   & 0.8070 & 0.4449 & 0.67  & 0.7904 & 0.3738 & 1.34  & 0.7945 & 0.1245 & 0.55 & 0.8609 & 0.1490 & 0.93  \\
		Wide\&Deep & 0.7984 & 0.4523 & 1.75  & 0.7928 & 0.3723 & 1.03  & 0.7916 & 0.1258 & 0.92 & 0.8643 & 0.1488 & 0.53  \\
		Deep\&Cross & 0.7987 & 0.4522 & 1.72  & 0.7935 & 0.3719 & 0.95  & 0.7947 & 0.1242 & 0.53 & \uline{0.8657} & 0.1472 & 0.37 \\
		NFM  & 0.8042 & 0.4469 & 1.02  & 0.7913 & 0.3730 & 1.23  & 0.7910 & 0.1256 & 1.00 & 0.8582 & 0.1506 & 1.25 \\
		DeepFM & 0.8036 & 0.4481 & 1.10  & 0.7939 & 0.3715 & 0.89  & 0.7917 & 0.1247 & 0.91 & 0.8626 & 0.1501 & 0.73 \\
		IPNN   & 0.8092 & 0.4420 & 0.40  & 0.7970 & 0.3698 & 0.50  & 0.7939 & 0.1240 & 0.63 & 0.8656 & \uline{0.1469} & 0.38 \\
		OPNN   & \uline{0.8102} & \uline{0.4417} & 0.27  & 0.7943 & 0.3715 & 0.84  & 0.7937 & 0.1242 & 0.66 & 0.8652 & 0.1480 & 0.43 \\
		xDeepFM & 0.8094 & 0.4421 & 0.37  & 0.7963 & 0.3707 & 0.59  & \uline{0.7950} & \uline{0.1239} & 0.49 & 0.8645 & 0.1481 & 0.51 \\
		AutoInt & 0.8082 & 0.4433 & 0.52  & 0.7929 & 0.3725 & 1.02  & 0.7909 & 0.1262 & 1.01 & 0.8632 & 0.1482 & 0.66 \\
		AutoGroup & 0.8089 & 0.4426 & 0.43  & \uline{0.7982} & \uline{0.3691} & 0.35  & 0.7949 & 0.1244 & 0.97 & 0.8637 & 0.1478 & 0.60 \\
		AutoFIS-FM & 0.8062 & 0.4452 & 0.77  & 0.7945 & 0.3712 & 0.82  & 0.7887 & 0.1268 & 1.29 & 0.8526 & 0.1533 & 1.91 \\
		AutoFIS-IPNN & 0.8094 & 0.4422 & 0.37  & 0.7978 & 0.3695 & 0.40  & 0.7945 & 0.1245 & 0.55 & 0.8644 & 0.1482 & 0.52 \\
		\midrule
		CELL (-DNA, Geno) & 0.8081 & 0.4436 & 0.53  & 0.7941 & 0.3713 & 0.87  & 0.7920 & 0.1253 & 0.87 & 0.8633 & 0.1486 & 0.65 \\
		CELL (-DNA)  & 0.8108 & 0.4408 & 0.20  & 0.7983 & 0.3689 & 0.34  & 0.7962 & 0.1234 & 0.34 & 0.8648 & 0.1483 & 0.47 \\
		CELL (-Geno)  & 0.8100  & 0.4415 & 0.30  & 0.7993 & 0.3682 & 0.21  & 0.7953 & 0.1238 & 0.45 & 0.8662 & 0.1472 & 0.31 \\
		\textbf{CELL}  & \textbf{0.8124}$^{\star}$ & \textbf{0.4393}$^{\star}$ & -     & \textbf{0.8010}$^{\star}$ & \textbf{0.3671}$^{\star}$ & -     & \textbf{0.7989}$^{\lozenge}$ & \textbf{0.1229}$^{\lozenge}$ & - & \textbf{0.8689}$^{\star}$ & \textbf{0.1460}$^{\star}$ & - \\
		\bottomrule
	\end{tabular}%
	\label{table02} 
\end{table*}%

\subsection{Experimental Settings}
\subsubsection{Evaluation Metrics} Two widely used metrics, AUC (Area Under Curve) and Logloss (cross-entropy loss) are selected to evaluate the recommendation performance.

\newcolumntype{L}[1]{>{\raggedright\arraybackslash}p{#1}}
\newcolumntype{C}[1]{>{\centering\arraybackslash}p{#1}}
\newcolumntype{R}[1]{>{\raggedleft\arraybackslash}p{#1}}

\begin{table}[!t]
	\centering
	\caption{Training cost of CELL (GPU hours).}
	\begin{tabular}{C{2cm}C{1.4cm}C{1.4cm}C{1.4cm}}
		\toprule
		Dataset & Stage I & Stage II & Stage III \\
		\midrule
		Criteo & $\sim$8.4 & $\sim$1.4 & $\sim$1.5 \\
		Avazu & $\sim$18.5 & $\sim$0.8 & $\sim$0.8 \\
		Huawei & $\sim$6.9 & $\sim$1.7 & $\sim$2.1 \\
		FinTech & $\sim$6.3 & $\sim$1.1 & $\sim$1.2 \\
		\bottomrule
	\end{tabular}%
	\label{table3} 
	\vspace{-0.5cm}
\end{table}%

\subsubsection{Baselines} The standard and state-of-the-art approaches we use as our baselines are: LR~\cite{richardson2007predicting}, FM~\cite{rendle2010fm}, AFM~\cite{xiao2017attentional}, FFM~\cite{juan2016field}, Wide\&Deep~\cite{cheng2016wide}, Deep\&Cross~\cite{wang2017deep}, NFM~\cite{he2017neural}, DeepFM~\cite{guo2017deepfm}, IPNN, OPNN~\cite{qu2016product}, xDeepFM~\cite{lian2018xdeepfm}, AutoInt~\cite{song2019autoint}, AutoGroup~\cite{liu2020autogroup}, AutoFIS-FM, and AutoFIS-IPNN~\cite{liu2020autofis}. We calculate the $p$-values for CELL and the best baseline by repeating the experiments 10 times by changing the random seeds. The two-tailed pairwise t-test is performed to detect the significant difference. We use $^{\star}$ and $^{\lozenge}$ to represent significance level $p$-value $<10^{-5}$ and $p$-value $<0.05$.

\subsubsection{Implementation Details}
Specifically, to implement our proposed CELL, in the first stage we use Adam~\cite{kingma2014adam} as the optimizer for DNA search with initial learning rate $\xi=10^{-3}$, momentum $(0.5, 0.999)$, weight decay $10^{-3}$.
In the second stage, we use RDA~\cite{xiao2009dual, chao2019generalization} as the optimizer for genome  search with the learning rate $\gamma=10^{-3}$, adjustable hyper-parameters $c=0.5, \mu=0.8$. We set the mutation mechanism as $\lambda=0.1$, $\sigma=5$, $\tau=100$.
In the model functioning stage, we set the depth of MLP as $2$ with $400$ neurons per layer.
In order to verify the effectiveness of DNA search and genome search, we design ablation studies by dropping one or two stages, i.e., CELL (-DNA) denotes CELL with DNA search dropped, which means operations are randomly assigned to each feature pair. CELL (-Geno) denotes CELL with genome search dropped, which means interactions are retained with equal relevance fitness. CELL (-DNA, Geno) combines the settings of CELL (-DNA) and CELL (-Geno).

\begin{figure*}[!t]
	\centering
	\includegraphics[width=\textwidth]{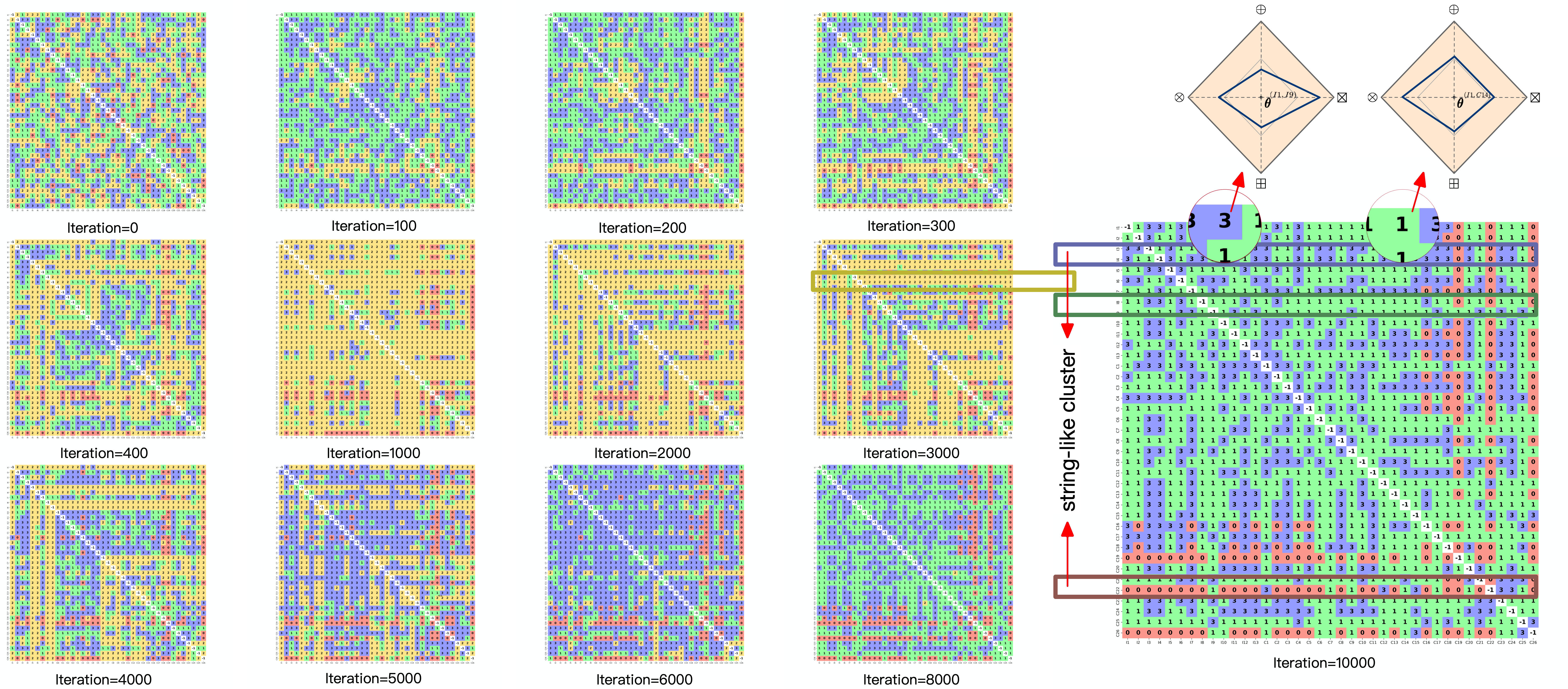}
	\caption{Visualization of DNA search on Criteo dataset and the demonstration of diagnosed fitness on operations.} \label{fig03}
	\vspace{-0.1cm}
\end{figure*} 

\begin{figure*}[!t]
	\centering
	\includegraphics[width=\textwidth]{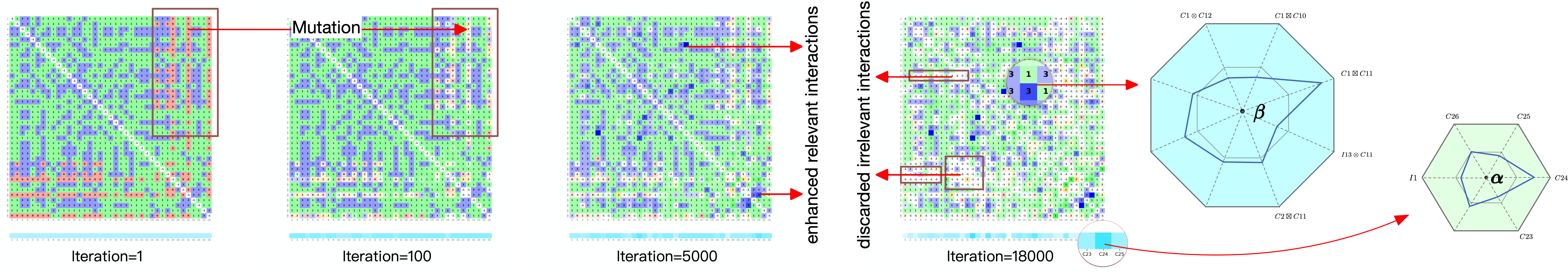}
	\caption{Visualization of genome search on Criteo dataset and the demonstration of diagnosed fitness on features and interactions.} \label{fig04}
	\vspace{-0.2cm}
\end{figure*} 

\subsection{Performance Comparison}

Table \ref{table02} reports the performance of compared baselines and CELL on the four datasets. Impr is the relative AUC improvement. It is worth mentioning that, unlike AUC, the smaller the Logloss value, the better the result. In practice, an improvement of 0.001-level in AUC or Logloss is usually regarded as being significant, because it will lead to a large increase in the company’s revenue due to a large user base, which has been pointed out in many existing articles~\cite{cheng2016wide, guo2017deepfm, yu2021xcrossnet,shao2021toward}.
From the experimental results, we have the following key observations: 

First, most neural network models outperform shallow models, i.e., LR, FM, which indicates that MLP can learn non-linear interactions and endow representation capabilities. Furthermore, OPNN, AutoGroup, and IPNN, the representatives of interactions modeled by element-wise outer and inner products, are the best baselines on Criteo, Avazu, and FinTech. Deep\&Cross and xDeepFM, the representatives of interactions modeled by cross networks and compressed interaction networks, are the best baselines on FinTech and Huawei. The results are further consistent with our statement that we cannot ensure which pre-designed operations are better because they are poorly adaptable to tasks and datasets.

Second, AUC and Logloss of CELL have significant improvement over previous baselines on all four datasets. This gain in performance can be attributed to adaptively modeling interactions by evolving to find suitable operations. Another benefit comes from that CELL can distinguish the relevance of features and interactions. Instead of numerating interactions in a brute-force way, CELL is able to select task-friendly features and interactions to avoid extra noise from irrelevant features and interactions. 

Third, from the results of ablation studies, we can observe that CELL outperforms CELL (-DNA), and CELL (-Geno) outperforms CELL (-DNA, Geno), which indicate that DNA search is effective because it can diagnose the fitness of the model on operations to obtain appropriate operations. Meanwhile, the superior performance of CELL over CELL (-Geno) and the superior performance of CELL (-DNA) over CELL (-DNA, Geno) indicate that genome search can diagnose the fitness of the model on features and interactions to find relevant ones and promote the model.

Beyond accuracy, we also display the time complexity of three stages of CELL in Table \ref{table3}. Note that every stage runs on a single GPU. The training cost of each stage is consistent with our analysis. Moreover, the entire training cost for each dataset is less than one day. CELL has excellent efficiency compared to previous automated machine learning approaches that usually run on massive GPUs for days~\cite{song2020towards}. On the Avazu dataset, the training cost of the first stage is a little longer, mainly because the dimension of embeddings is set relatively higher than the other datasets.

\begin{figure*}[!t]
	\centering
	\subfloat[]{
		\begin{minipage}[t]{0.4\linewidth}
			\centering
			\includegraphics[width=7cm]{synthetic_new_a02.pdf}
		\end{minipage} \label{fig5a}
	}%
	\subfloat[]{
		\begin{minipage}[t]{0.3\linewidth}
			\centering
			\includegraphics[width=4.5cm]{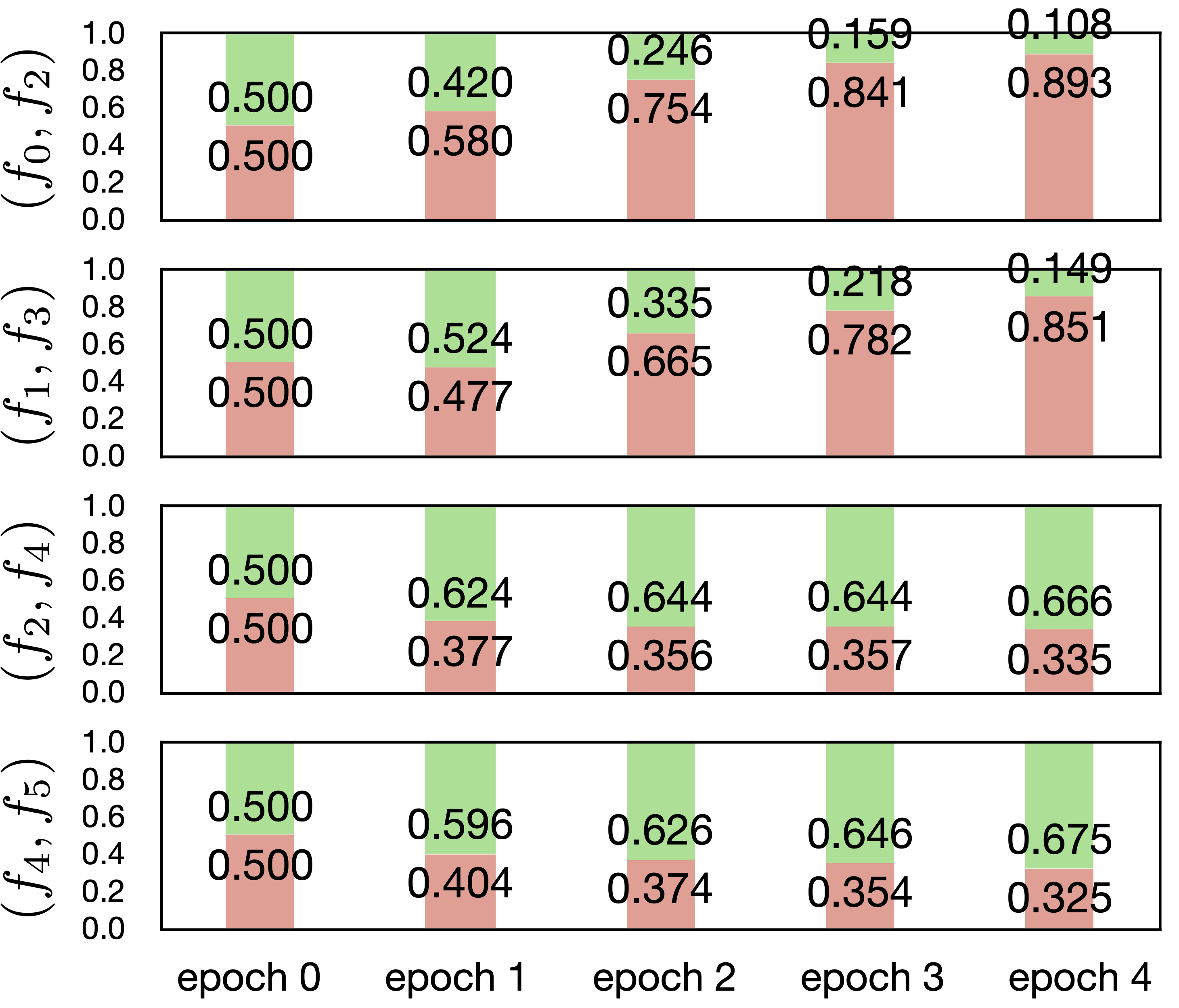}
		\end{minipage}\label{fig5b}
	}%
	\subfloat[]{
		\begin{minipage}[t]{0.3\linewidth}
			\flushleft
			\includegraphics[width=4.5cm]{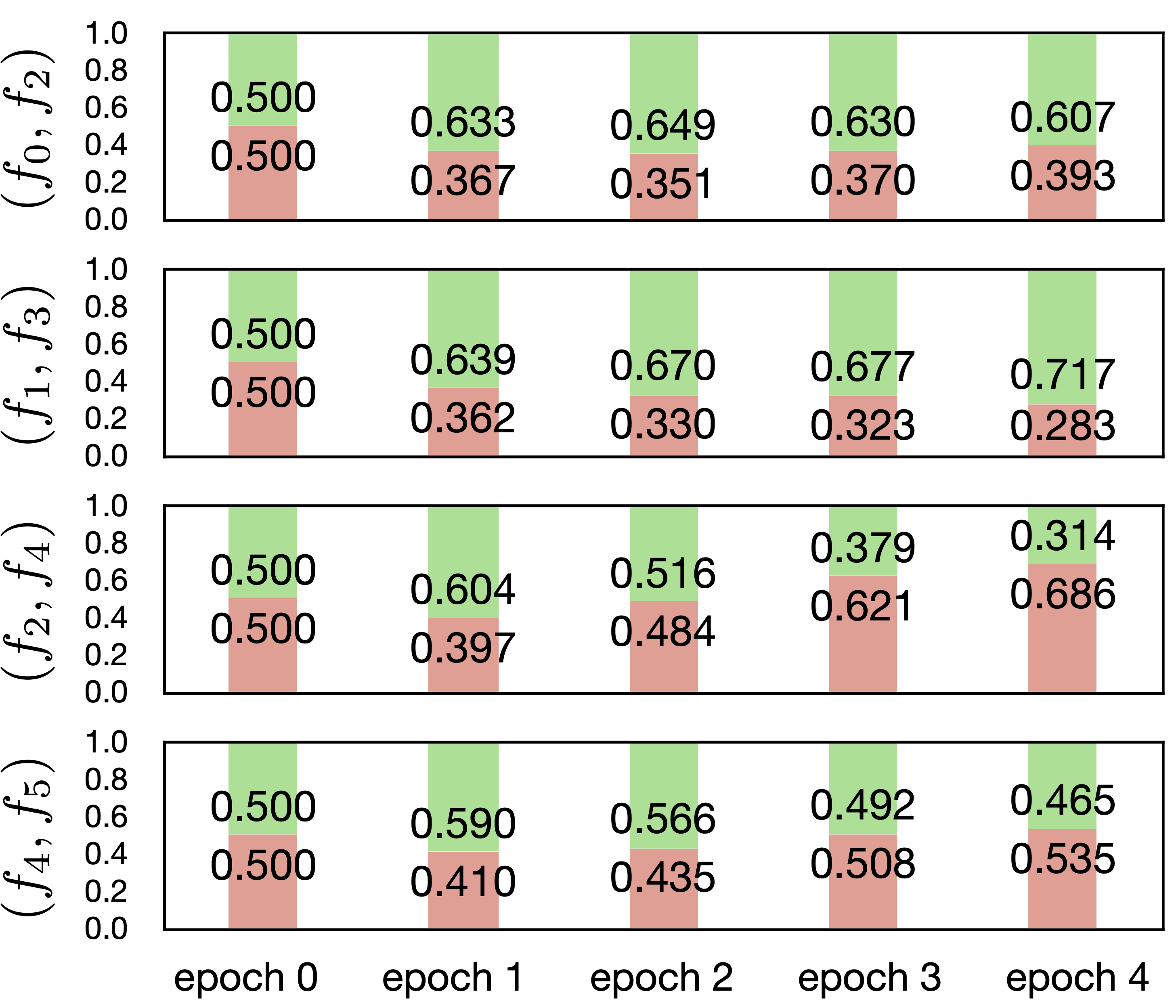}
		\end{minipage}\label{fig5c}
	}%
	
	\centering
	\caption{ Results of the synthetic experimental studies. (a) Diagnosed relevance fitness of interactions. (b) Diagnosed fitness on operations to model the interactions associated with the selected poly-2 terms. (c) Diagnosed fitness on the switched dataset. }
	\vspace{-0.3cm}
\end{figure*}
\label{fig5}

\subsection{Visualization of Evolution Path}

To make it clear how the model evolves to select suitable operations, features, and interactions under the task guidance, we visualize the evolution path of DNA search and genome search on the Criteo dataset, which contains 13 integer feature fields ``$I1 \sim I13$" and 26 categorical feature fields ``$C1 \sim C26$". If we use the following encoding $\oplus=0, \otimes=1,  \boxplus=2, \boxtimes=3$, the fitness diagnosis results can be represented as a matrix. Furthermore, we assign different colors for operations to describe a \textbf{gene map} of the model, where each gene indicates an interaction, i.e., red ``$0$", green ``$1$", yellow ``$2$", blue ``$3$". For example, green ``$1$" in the block ``$I1\times C12$" means that element-wise product~$\otimes$ is diagnosed as the fittest operation for feature $I1$ to interact with feature $C12$. In genome search, in order to express the relevance of features and interactions, we highlight some genes according to the relevance fitness parameters. The darker color represents the feature or the interaction that is diagnosed as the more relevant one, and the lighter color represents the less relevant one.

We can observe that, in Fig. \ref{fig03}, operations were randomly assigned to model all interactions at the beginning. Then, the gene map evolved rapidly in the early iterations, and was occupied by a large number of operations $\otimes$ and $\boxtimes$, which could be regarded as the model falling into the first local optimal. After more batches of data were learned, the gene map changed drastically. In the 1000th iteration, the gene map was significantly occupied by operations $\boxplus$, which could be regarded as the model falling into the second local optimal. What interests us is that the subsequent evolution direction has undergone a dramatic change. We can see more operations $\boxplus$ have been replaced by operations $\otimes$ and $\boxtimes$ when the model converges. In addition to these dynamic changes, there are some static patterns. For example, operations had the effect of string-like clustering on the gene map as we marked, which means that interactions containing the same feature were more likely to be modeled by the same operation.

In Fig. \ref{fig04}, all features and interactions shared equal relevance fitness at the beginning. Later, some were discovered as relevant features and interactions (the color becomes darker), while most of the others became less important (the color becomes lighter), and some mutated into new interactions. The relevance fitness parameters of a part of features and interactions were reduced and truncated to $0$. We discarded these irrelevant features and interactions, so their genes became white ``$-1$". Another interesting observation is that, in DNA search, the operations of some interactions hardly changed with the increase of the learning batches. Combined with genome search, we can find that these interactions are almost either highly relevant or irrelevant interactions.

\subsection{Synthetic Experiments} 

To further validate the effectiveness of the fitness diagnosis technique of CELL, we conduct a synthetic experiment. We artificially construct such a synthetic dataset, where the input feature $\pmb{f}$ of this dataset is sampled from the Criteo dataset containing $N$ categories of $m$ fields, and the output $y$ is binary labeled and generated from a poly-2 function. Based on this synthetic dataset, we investigate whether the fitness diagnosis technique can diagnose to find (1) the relevant interactions and (2)~the operations to model these interactions.

Here we choose $m=6, N=4481$ to test the effectiveness of our model. We use the element-wise sum $\oplus$ and the element-wise product $\otimes$ to build such sums of poly-2 terms. The output binary label $y$ is given as: 
\begin{equation} \small
y= \delta\left( \sum_{i,j\in C_1} f_i\oplus f_j + \sum_{i,j\in C_2} f_i \otimes f_j +\epsilon \right), 
\end{equation}
\begin{equation} \small
\begin{aligned}
& \delta(q) = \begin{cases}
1 ,& \textbf{if} \quad q \geqslant \text{threshold}, \\
0 ,\;\;\; & \textbf{otherwise}.
\end{cases}
\end{aligned} \label{eq3}
\end{equation} 

The feature pairs are $i.i.d.$ sampled to build the train and test datasets. We also add a small random noise $\epsilon$ to the sampled data. 
First, the selected poly-2 term sets $C_1=\{ (f_0,f_2), (f_1,f_3) \}$ and $C_2=\{ (f_2,f_4), (f_4,f_5) \}$ are initialized and fixed. 
After effortless training, CELL achieved AUC$=0.9920$, Logloss$=0.1269$ on this synthetic dataset, which means CELL can consistently discover the pre-defined interaction patterns for interacting feature pairs and make the right predictions. 
Figure~\ref{fig5b} shows the Diagnosed fitness on operations to model the interactions associated with the selected poly-2 terms. Red color indicates element-wise sum $\oplus$ and green color indicates element-wise product $\otimes$. In order to further validate the diagnosis results are not affected by assigned features, we switch the operations for the selected poly-2 terms to generate another new dataset. Thus, this time the selected poly-2 terms $C_1=\{ (f_2,f_4), (f_4,f_5) \}$ and $C_2=\{ (f_0,f_2), (f_1,f_3) \}$ are initialized and fixed. Figure~\ref{fig5c} shows the diagnosed fitness on this switched dataset.

As shown in Figure~\ref{fig5a}, four selected poly-2 terms have been clearly extracted through fitness diagnosis, indicating that CELL can precisely discover the relevant interactions. Furthermore, in Figure~\ref{fig5b}, the correct operations for modeling the relevant interactions are diagnosed as better learning proficiency. Moreover, in Figure~\ref{fig5c}, as switching the operations, the diagnosed fitness is concomitantly reversed, which confirms that our diagnosis technique on operations is effective and independent of assigned features.

\begin{figure*}[!t]
	\centering
	\includegraphics[width=\textwidth]{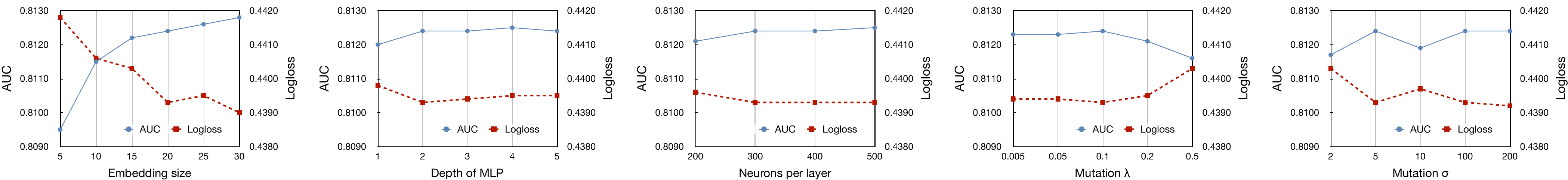}
	\caption{Impact of hyper-parameters on Criteo dataset.} \label{fig05}
	\vspace{-0.25cm}
\end{figure*} 
\begin{figure*}[!t]
	\centering
	\includegraphics[width=\textwidth]{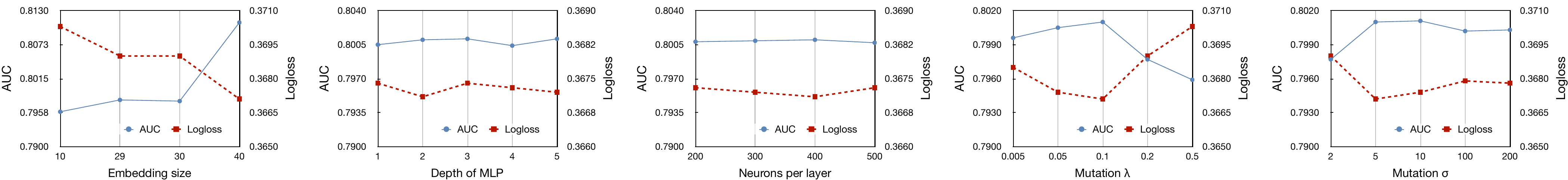}
	\caption{Impact of hyper-parameters on Avazu dataset.} \label{fig06}
	\vspace{-0.25cm}
\end{figure*} 
\begin{figure*}[!t]
	\centering
	\includegraphics[width=\textwidth]{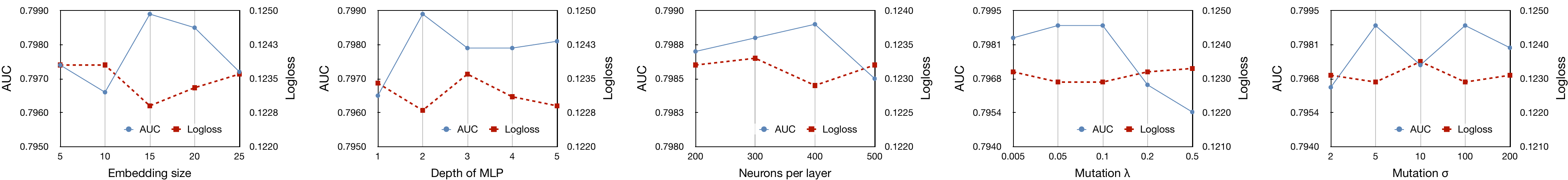}
	\caption{Impact of hyper-parameters on Huawei dataset.} \label{fig07}
	\vspace{-0.25cm}
\end{figure*} 
\begin{figure*}[t]
	\centering
	\includegraphics[width=\textwidth]{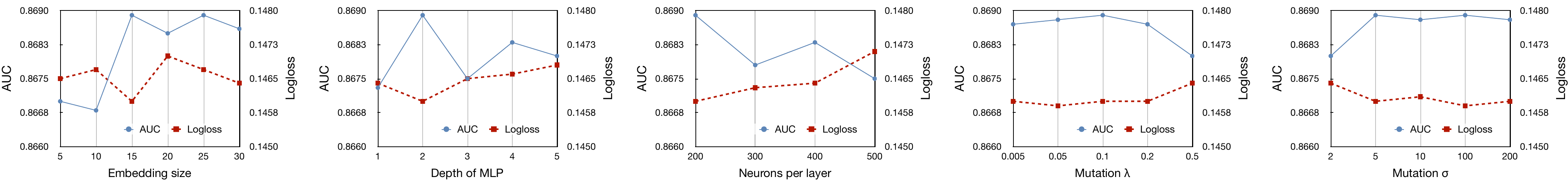}
	\caption{Impact of hyper-parameters on FinTech dataset.} \label{fig08}
	\vspace{-0.35cm}
\end{figure*} 

\subsection{Hyper-parameter Studies}

We investigate the impact of hyper-parameters of CELL, including (1) embedding size; (2) depth of MLP; (3) neurons per layer; (4) mutation threshold $\lambda$; (5) mutation probability parameter $\sigma$. The experimental results on four datasets in terms of AUC and Logloss are shown from Fig. \ref{fig05} to Fig. \ref{fig08}.

From the experimental results, we can observe that embedding size has the most significant impact on performance. Meanwhile, even with small embedding sizes, CELL still has comparable performance to some popular baselines with large embedding sizes, e.g., for the advertising dataset Criteo, CELL achieves AUC $>0.8090$ and Logloss $<0.4420$ with the embedding size set as $5$, which is better than DeepFM with an embedding size set as $20$. For the financial dataset FinTech, we get a similar observation that CELL achieves AUC $>0.8660$ and Logloss $<0.1470$ with the embedding size set as $5$, which is better than Deep\&Cross with embedding size set as $15$. Note that Deep\&Cross is regarded as one of the best baselines on the FinTech dataset. Furthermore, the model performance boosts with the depth of MLP at the beginning. However, it barely boosts when the depth of MLP is set to greater than $2$. Also, the model performance barely boosts as the number of neurons per layer increases from $200$ to $500$. We consider $400$ is a more suitable setting to avoid the model overfitting. Moreover, the mutation threshold $\lambda$ should not be set too large, and the mutation probability parameter $\sigma$ should not be set too small, both to avoid over-random mutation.

\section{Conclusion} \label{section7}

This paper presents a nature-inspired learning approach to select feature interactions for recommender systems, namely Cognitive Evolutionary Learning (CELL). CELL can adaptively evolve a model to find (1) the proper operation for modeling each interaction and (2) relevant features and interactions to the task. 
Attributed to the proposed fitness diagnosis technique, CELL can diagnose the learning abilities of inside components of models during training, thereby enhancing interpretability and interpreting the mechanism of interaction modeling and selection. We conducted extensive experiments on four datasets, including three publicly available advertising datasets and one collected financial dataset. Experimental results have proved that CELL can significantly outperform state-of-the-art approaches. By tracing the evolution path, we can reveal the learning abilities of models during training and interpret how to select operations, features, and interactions. The synthetic experiments are further consistent with our statement because CELL can clearly extract the pre-defined interaction patterns. The CELL framework of utilizing evolutionary learning and diagnosis techniques is new thinking that helps us build a model under task guidance and is not limited to the instantiation in this paper. We encourage more task-oriented instantiations to be proposed based on our work.

\normalem
\bibliographystyle{IEEEtran}
\bibliography{anonymous-submission-latex-2023}

\begin{thebibliography}{10}
\providecommand{\url}[1]{#1}
\csname url@samestyle\endcsname
\providecommand{\newblock}{\relax}
\providecommand{\bibinfo}[2]{#2}
\providecommand{\BIBentrySTDinterwordspacing}{\spaceskip=0pt\relax}
\providecommand{\BIBentryALTinterwordstretchfactor}{4}
\providecommand{\BIBentryALTinterwordspacing}{\spaceskip=\fontdimen2\font plus
\BIBentryALTinterwordstretchfactor\fontdimen3\font minus
  \fontdimen4\font\relax}
\providecommand{\BIBforeignlanguage}[2]{{%
\expandafter\ifx\csname l@#1\endcsname\relax
\typeout{** WARNING: IEEEtran.bst: No hyphenation pattern has been}%
\typeout{** loaded for the language `#1'. Using the pattern for}%
\typeout{** the default language instead.}%
\else
\language=\csname l@#1\endcsname
\fi
#2}}
\providecommand{\BIBdecl}{\relax}
\BIBdecl

\bibitem{guo2021dual}
W.~Guo \emph{et~al.}, ``Dual graph enhanced embedding neural network for ctr
  prediction,'' in \emph{KDD}, 2021, pp. 496--504.

\bibitem{li2021dual}
P.~Li \emph{et~al.}, ``Dual attentive sequential learning for cross-domain
  click-through rate prediction,'' in \emph{KDD}, 2021, pp. 3172--3180.

\bibitem{shi2020deep}
S.-T. Shi, W.~Zheng, J.~Tang, Q.-G. Chen, Y.~Hu, J.~Zhu, and M.~Li, ``Deep
  time-stream framework for click-through rate prediction by tracking interest
  evolution,'' in \emph{AAAI}, 2020, pp. 5726--5733.

\bibitem{lyu2020deep}
Z.~Lyu, Y.~Dong, C.~Huo, and W.~Ren, ``Deep match to rank model for
  personalized click-through rate prediction,'' in \emph{AAAI}, vol.~34,
  no.~01, 2020, pp. 156--163.

\bibitem{shao2021toward}
Q.~Shao \emph{et~al.}, ``Toward intelligent financial advisors for identifying
  potential clients: A multitask perspective,'' \emph{Big Data Mining and
  Analytics}, vol.~5, no.~1, pp. 64--78, 2021.

\bibitem{yu2021xcrossnet}
R.~Yu \emph{et~al.}, ``Xcrossnet: Feature structure-oriented learning for
  click-through rate prediction,'' in \emph{PAKDD}, 2021, pp. 436--447.

\bibitem{dash1997feature}
M.~Dash and H.~Liu, ``Feature selection for classification,'' \emph{Intelligent
  data analysis}, vol.~1, no. 1-4, pp. 131--156, 1997.

\bibitem{li2022survey}
X.~Li, Y.~Wang, and R.~Ruiz, ``A survey on sparse learning models for feature
  selection,'' \emph{IEEE transactions on Cybernetics}, pp. 1642--1660, 2022.

\bibitem{rendle2010fm}
S.~Rendle, ``Factorization machines,'' in \emph{ICDM}.\hskip 1em plus 0.5em
  minus 0.4em\relax IEEE, 2010, pp. 995--1000.

\bibitem{he2017neural}
X.~He and T.-S. Chua, ``Neural factorization machines for sparse predictive
  analytics,'' in \emph{SIGIR}, 2017, pp. 355--364.

\bibitem{xiao2017attentional}
J.~Xiao \emph{et~al.}, ``Attentional factorization machines: learning the
  weight of feature interactions via attention networks,'' in \emph{IJCAI},
  2017, pp. 3119--3125.

\bibitem{guo2017deepfm}
H.~Guo \emph{et~al.}, ``Deepfm: {A} factorization-machine based neural network
  for {CTR} prediction,'' in \emph{IJCAI}, 2017, pp. 1725--1731.

\bibitem{xue2015survey}
B.~Xue, M.~Zhang, W.~N. Browne, and X.~Yao, ``A survey on evolutionary
  computation approaches to feature selection,'' \emph{IEEE Transactions on
  Evolutionary Computation}, vol.~20, no.~4, pp. 606--626, 2015.

\bibitem{zhou2019evolutionary}
Z.-H. Zhou, Y.~Yu, and C.~Qian, \emph{Evolutionary learning: Advances in
  theories and algorithms}.\hskip 1em plus 0.5em minus 0.4em\relax Springer,
  2019.

\bibitem{yu2023cognitive}
R.~Yu, X.~Xu, Y.~Ye, Q.~Liu, and E.~Chen, ``Cognitive evolutionary search to
  select feature interactions for click-through rate prediction,'' in
  \emph{KDD}, 2023, pp. 3151--3161.

\bibitem{telikani2021evolutionary}
A.~Telikani, A.~Tahmassebi, W.~Banzhaf, and A.~H. Gandomi, ``Evolutionary
  machine learning: A survey,'' \emph{ACM Computing Surveys (CSUR)}, vol.~54,
  no.~8, pp. 1--35, 2021.

\bibitem{tran2017new}
B.~Tran, B.~Xue, and M.~Zhang, ``A new representation in pso for
  discretization-based feature selection,'' \emph{IEEE Transactions on
  Cybernetics}, vol.~48, no.~6, pp. 1733--1746, 2017.

\bibitem{cheng2021steering}
F.~Cheng, F.~Chu, Y.~Xu, and L.~Zhang, ``A steering-matrix-based multiobjective
  evolutionary algorithm for high-dimensional feature selection,'' \emph{IEEE
  Transactions on Cybernetics}, vol.~52, no.~9, pp. 9695--9708, 2022.

\bibitem{liu2012feature}
H.~Liu and H.~Motoda, \emph{Feature selection for knowledge discovery and data
  mining}.\hskip 1em plus 0.5em minus 0.4em\relax Springer Science \& Business
  Media, 2012, vol. 454.

\bibitem{zhou2019deep}
G.~Zhou, N.~Mou, Y.~Fan, Q.~Pi, W.~Bian, C.~Zhou, X.~Zhu, and K.~Gai, ``Deep
  interest evolution network for click-through rate prediction,'' in
  \emph{AAAI}, 2019, pp. 5941--5948.

\bibitem{xie2021fives}
Y.~Xie \emph{et~al.}, ``Fives: Feature interaction via edge search for
  large-scale tabular data,'' in \emph{KDD}, 2021, pp. 3795--3805.

\bibitem{guo2021embedding}
H.~Guo, B.~Chen \emph{et~al.}, ``An embedding learning framework for numerical
  features in ctr prediction,'' in \emph{KDD}, 2021, pp. 2910--2918.

\bibitem{juan2016field}
Y.~Juan, Y.~Zhuang, W.-S. Chin, and C.-J. Lin, ``Field-aware factorization
  machines for ctr prediction,'' in \emph{RecSys}, 2016, pp. 43--50.

\bibitem{zhang2016deep}
W.~Zhang, T.~Du, and J.~Wang, ``Deep learning over multi-field categorical
  data,'' in \emph{European Conference on Information Retrieval}.\hskip 1em
  plus 0.5em minus 0.4em\relax Springer, 2016, pp. 45--57.

\bibitem{qu2016product}
Y.~Qu \emph{et~al.}, ``Product-based neural networks for user response
  prediction,'' in \emph{ICDM}.\hskip 1em plus 0.5em minus 0.4em\relax IEEE,
  2016, pp. 1149--1154.

\bibitem{qu2018product}
Y.~Qu, B.~Fang \emph{et~al.}, ``Product-based neural networks for user response
  prediction over multi-field categorical data,'' \emph{ACM Transactions on
  Information Systems (TOIS)}, vol.~37, no.~1, pp. 1--35, 2018.

\bibitem{lian2018xdeepfm}
J.~Lian \emph{et~al.}, ``xdeepfm: Combining explicit and implicit feature
  interactions for recommender systems,'' in \emph{KDD}, 2018, pp. 1754--1763.

\bibitem{cheng2016wide}
H.-T. Cheng, L.~Koc, J.~Harmsen, T.~Shaked \emph{et~al.}, ``Wide \& deep
  learning for recommender systems,'' in \emph{Proceedings of the 1st Workshop
  on Deep Learning for Recommender Systems}, 2016, pp. 7--10.

\bibitem{wang2017deep}
R.~Wang, B.~Fu, G.~Fu, and M.~Wang, ``Deep \& cross network for ad click
  predictions,'' in \emph{ADKDD}, 2017, pp. 1--7.

\bibitem{back1996evolutionary}
T.~Back, \emph{Evolutionary algorithms in theory and practice: evolution
  strategies, evolutionary programming, genetic algorithms}.\hskip 1em plus
  0.5em minus 0.4em\relax Oxford University Press, 1996.

\bibitem{juang2021navigation}
C.-F. Juang, C.-Y. Chou, and C.-T. Lin, ``Navigation of a fuzzy-controlled
  wheeled robot through the combination of expert knowledge and data-driven
  multiobjective evolutionary learning,'' \emph{IEEE Transactions on
  Cybernetics}, vol.~52, no.~8, pp. 7388--7401, 2022.

\bibitem{zheng2020reconstruction}
X.~Zheng, W.~Wu, W.~Deng, C.~Yang, and K.~Huang, ``Reconstruction of tree
  network via evolutionary game data analysis,'' \emph{IEEE Transactions on
  Cybernetics}, vol.~52, no.~7, pp. 6083--6094, 2022.

\bibitem{chen2020evolutionary}
K.~Chen, B.~Xue, M.~Zhang, and F.~Zhou, ``An evolutionary multitasking-based
  feature selection method for high-dimensional classification,'' \emph{IEEE
  Transactions on Cybernetics}, vol.~52, no.~7, pp. 7172--7186, 2022.

\bibitem{song2021fast}
X.-F. Song, Y.~Zhang, D.-W. Gong, and X.-Z. Gao, ``A fast hybrid feature
  selection based on correlation-guided clustering and particle swarm
  optimization for high-dimensional data,'' \emph{IEEE Transactions on
  Cybernetics}, vol.~52, no.~9, pp. 9573--9586, 2022.

\bibitem{khawar2020autofeature}
F.~Khawar \emph{et~al.}, ``Autofeature: Searching for feature interactions and
  their architectures for click-through rate prediction,'' in \emph{CIKM},
  2020, pp. 625--634.

\bibitem{song2020towards}
Q.~Song, D.~Cheng \emph{et~al.}, ``Towards automated neural interaction
  discovery for click-through rate prediction,'' in \emph{KDD}, 2020, pp.
  945--955.

\bibitem{liu2020autogroup}
B.~Liu \emph{et~al.}, ``Autogroup: Automatic feature grouping for modelling
  explicit high-order feature interactions in ctr prediction,'' in
  \emph{SIGIR}, 2020, pp. 199--208.

\bibitem{dibello200631a}
L.~V. DiBello, L.~A. Roussos, and W.~Stout, ``A review of cognitively
  diagnostic assessment and a summary of psychometric models,'' \emph{Handbook
  of statistics}, vol.~26, pp. 979--1030, 2006.

\bibitem{de2011generalized}
J.~De~La~Torre, ``The generalized dina model framework,'' \emph{Psychometrika},
  vol.~76, no.~2, pp. 179--199, 2011.

\bibitem{tong2021item}
S.~Tong \emph{et~al.}, ``Item response ranking for cognitive diagnosis,'' in
  \emph{IJCAI}, 2021, pp. 1750--1756.

\bibitem{yue2021circumstances}
L.~Yue \emph{et~al.}, ``Circumstances enhanced criminal court view
  generation,'' in \emph{SIGIR}, 2021, pp. 1855--1859.

\bibitem{gu2021neuralac}
Y.~Gu \emph{et~al.}, ``Neuralac: Learning cooperation and competition effects
  for match outcome prediction,'' in \emph{AAAI}, vol.~35, no.~5, 2021, pp.
  4072--4080.

\bibitem{tao2022sminet}
W.~Tao, Y.~Li, L.~Li, Z.~Chen, H.~Wen, P.~Chen, T.~Liang, and Q.~Lu, ``Sminet:
  State-aware multi-aspect interests representation network for cold-start
  users recommendation,'' in \emph{AAAI}, vol.~36, no.~8, 2022, pp. 8476--8484.

\bibitem{liu2019darts}
H.~Liu, K.~Simonyan, and Y.~Yang, ``Darts: Differentiable architecture
  search,'' in \emph{ICLR}, 2019.

\bibitem{finn2017model}
C.~Finn, P.~Abbeel, and S.~Levine, ``Model-agnostic meta-learning for fast
  adaptation of deep networks,'' in \emph{ICML}.\hskip 1em plus 0.5em minus
  0.4em\relax PMLR, 2017, pp. 1126--1135.

\bibitem{luketina2016scalable}
J.~Luketina, M.~Berglund, K.~Greff, and T.~Raiko, ``Scalable gradient-based
  tuning of continuous regularization hyperparameters,'' in \emph{ICML}.\hskip
  1em plus 0.5em minus 0.4em\relax PMLR, 2016, pp. 2952--2960.

\bibitem{metz2016unrolled}
L.~Metz, B.~Poole, D.~Pfau, and J.~Sohl-Dickstein, ``Unrolled generative
  adversarial networks,'' \emph{arXiv preprint arXiv:1611.02163}, 2016.

\bibitem{kingma2014adam}
D.~P. Kingma and J.~Ba, ``Adam: A method for stochastic optimization,''
  \emph{arXiv preprint arXiv:1412.6980}, 2014.

\bibitem{xiao2009dual}
L.~Xiao, ``Dual averaging method for regularized stochastic learning and online
  optimization,'' \emph{NeurIPS}, vol.~22, pp. 2116--2124, 2009.

\bibitem{chao2019generalization}
S.-K. Chao and G.~Cheng, ``A generalization of regularized dual averaging and
  its dynamics,'' \emph{arXiv preprint arXiv:1909.10072}, 2019.

\bibitem{liu2020autofis}
B.~Liu \emph{et~al.}, ``Autofis: Automatic feature interaction selection in
  factorization models for click-through rate prediction,'' in \emph{KDD},
  2020, pp. 2636--2645.

\bibitem{domingos2015master}
P.~Domingos, \emph{The master algorithm: How the quest for the ultimate
  learning machine will remake our world}.\hskip 1em plus 0.5em minus
  0.4em\relax Basic Books, 2015.

\bibitem{richardson2007predicting}
M.~Richardson \emph{et~al.}, ``Predicting clicks: estimating the click-through
  rate for new ads,'' in \emph{WWW}, 2007, pp. 521--530.

\bibitem{song2019autoint}
W.~Song \emph{et~al.}, ``Autoint: Automatic feature interaction learning via
  self-attentive neural networks,'' in \emph{CIKM}.\hskip 1em plus 0.5em minus
  0.4em\relax ACM, 2019, pp. 1161--1170.

\end{thebibliography}

\begin{IEEEbiography}[{\includegraphics[width=1in,height=1.25in,clip,keepaspectratio]{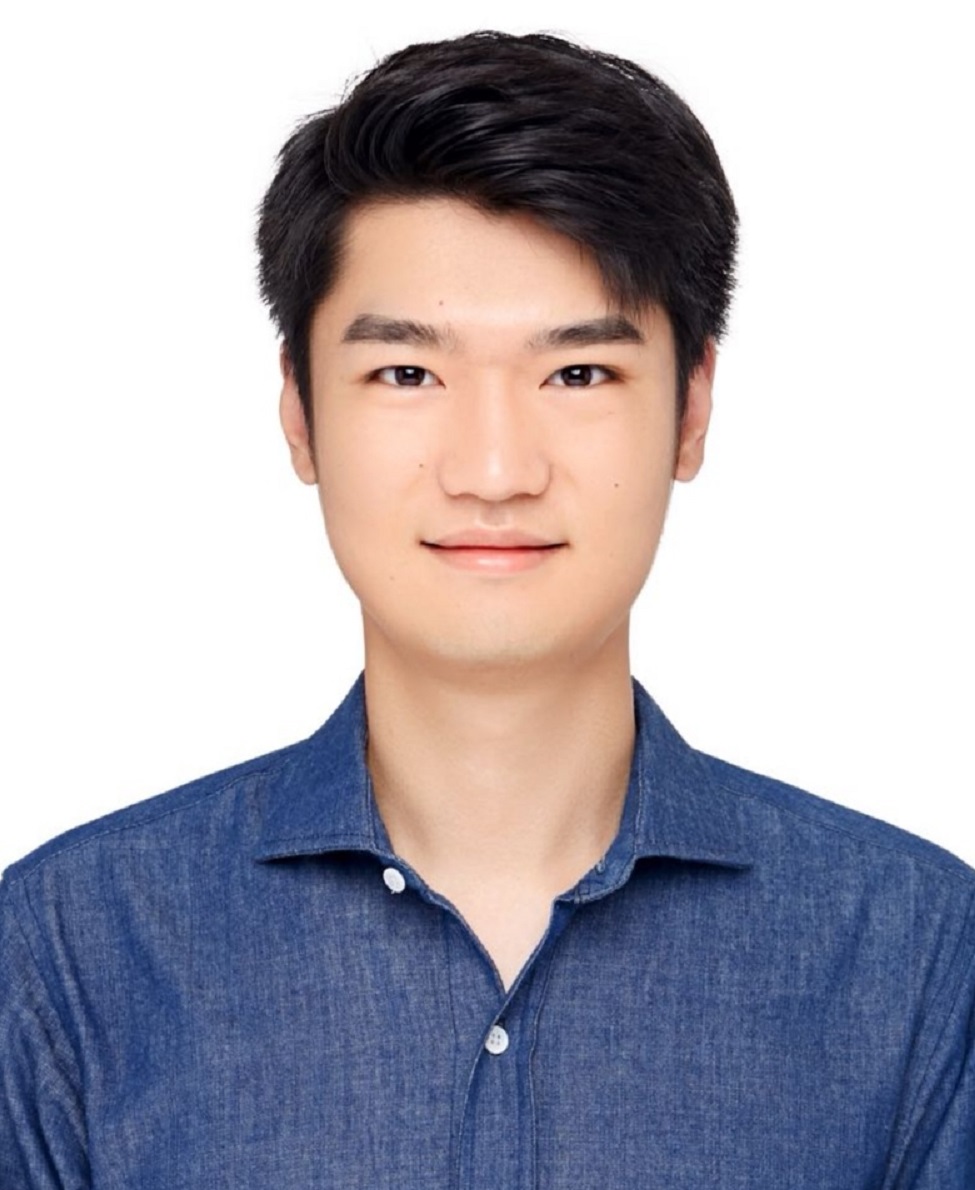}}]
	{Runlong Yu}
	received his Ph.D. degree in computer science and technology at the University of Science and Technology of China (USTC) in Hefei, China, in 2023, following his B.Eng. degree in computer science and technology from USTC, in 2017. He is currently a Postdoctoral Associate in the Department of Computer Science at the University of Pittsburgh. His research interests include data mining, machine learning, evolutionary computation, AI for science, and recommender systems. He has published over twenty papers in refereed journals and conference proceedings including IEEE TKDE, KAIS, ACM SIGKDD, ACM CIKM, ACM SIGIR, PAKDD, IEEE ICDM, IJCAI, AAAI, and DASFAA. He was the recipient of the KDD CUP 2019 PaddlePaddle Special Award and the 2020 CCF BDCI Contest championship.
\end{IEEEbiography}

\begin{IEEEbiography}[{\includegraphics[width=1in,height=1.25in,clip,keepaspectratio]{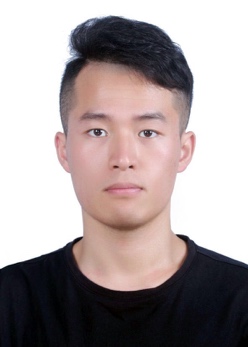}}]
	{Qixiang Shao} received his Master degree in computer science and technology at the University of Science and Technology of China (USTC) in Hefei, China, in 2023. Before that, he received his B.Eng. degree in software engineering from Nanjing University of Information Science and Technology (NUIST), Nanjing, China, in 2017. He is currently working as a software engineer at iFlytek. His research interests include recommender systems, data mining, and FinTech. 
\end{IEEEbiography}

\begin{IEEEbiography}[{\includegraphics[width=1in,height=1.25in,clip,keepaspectratio]{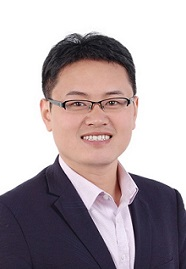}}]
	{Qi Liu}
	received the Ph.D. degree from University of Science and Technology of China (USTC), Hefei, China, in 2013. He is currently a Professor with Anhui Province Key Laboratory of Big Data Analysis and Application (BDAA), State Key Laboratory of Cognitive Intelligence, School of Computer Science and Technology, University of Science and Technology of China, Hefei, China. His research interests include data mining, machine learning, and recommender systems. He was the recipient of KDD' 18 (Research Track) Best Student Paper Award and ICDM' 11 Best Research Paper Award. He was the winner of the National Science Fund for Outstanding Youth Science Foundation in 2019. 
\end{IEEEbiography}

\begin{IEEEbiography}[{\includegraphics[width=1in,height=1.25in,clip,keepaspectratio]{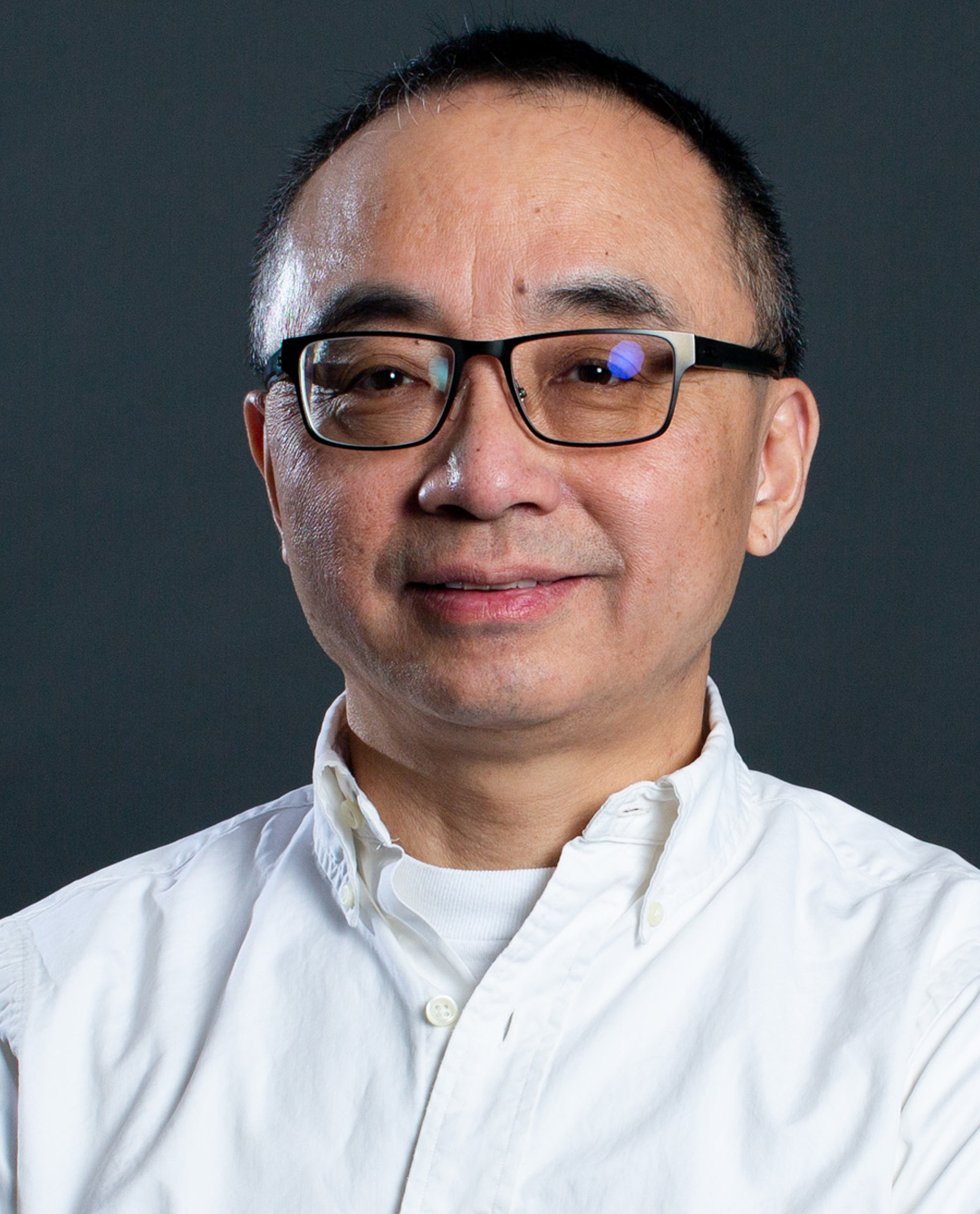}}]
	{Huan Liu} (F' 12) received the B.Eng. degree in computer science and electrical engineering from Shanghai JiaoTong University (SJTU), Shanghai, China, in 1983, and the Ph.D. degree in computer science from University of Southern California (USC), Los Angeles, CA, USA, in 1989. He is currently an Ira A. Fulton Professor of Computer Science and Engineering at Arizona State University (ASU), Tempe, AZ, USA. He is a Fellow of IEEE, ACM, AAAI, and AAAS.  His well-cited publications include books, book chapters, encyclopedia entries, and conference and journal papers. His research interests include data mining, machine learning, social computing, and feature selection.  He was recognized for excellence in teaching and research in computer science and engineering at Arizona State University. He serves on journal editorial boards and numerous conference program committees. He is also a Founding Organizer of the International Conference Series on Social Computing, Behavioral Cultural Modeling, and Prediction.
	
\end{IEEEbiography}

\begin{IEEEbiography}[{\includegraphics[width=1in,height=1.25in,clip,keepaspectratio]{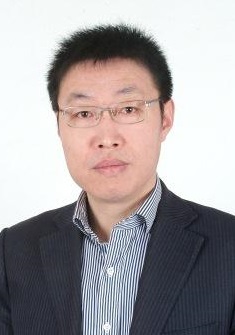}}]
	{Enhong Chen} (F' 24) received the Ph.D. degree from University of Science and Technology of China (USTC), Hefei, China, in 1996. 
	He is currently a Director (Professor) of Anhui Province Key Laboratory of Big Data Analysis and Application (BDAA), and a vice Director of State Key Laboratory of Cognitive Intelligence, University of Science and Technology of China, Hefei, China. He is an executive dean of School of Data Science of USTC. He is an IEEE Fellow, ACM Distinguished Member, and CCF Fellow.  His research interests include data mining, machine learning, and recommender systems. He has published 200+ refereed international conference and journal papers. He was the recipient of KDD' 08 Best Application Paper Award and ICDM' 11 Best Research Paper Award. He was the winner of the National Science Fund for Distinguished Young Scholars in 2013, scientific and technological innovation leading talent of `Ten Thousand Talent Program' in 2017.
\end{IEEEbiography}

\vfill

\end{document}